\documentclass[10pt,twocolumn,letterpaper]{article}

\usepackage[pagenumbers]{cvpr} 

\usepackage{graphicx}
\usepackage{amsmath}
\usepackage{amssymb}
\usepackage{booktabs}
\usepackage{csquotes}
\usepackage{multirow}
\usepackage{nicefrac}
\usepackage{xifthen}
\usepackage{lipsum}
\usepackage{dsfont}
\usepackage{color, colortbl}
\usepackage{enumitem}
\usepackage{float}

\usepackage[accsupp]{axessibility}  
\usepackage{balance}
\usepackage[symbol]{footmisc}

\usepackage{etoolbox}
\usepackage{silence}
\makeatletter
\robustify\@latex@warning@no@line
\makeatother
\usepackage{authblk}
\makeatletter
\renewcommand\AB@affilsepx{ \hspace{0.5em} \protect\Affilfont}

\makeatother

\usepackage[pagebackref,breaklinks,colorlinks]{hyperref}

\DeclareMathOperator*{\smax}{smax}

\usepackage[capitalize]{cleveref}
\crefname{section}{Sec.}{Secs.}
\Crefname{section}{Section}{Sections}
\Crefname{table}{Table}{Tables}
\crefname{table}{Tab.}{Tabs.}

\def\cvprPaperID{10905} 
\def\confName{CVPR}
\def\confYear{2022}

\newcommand{\pgd}{%
	$\text{\textsc{pgd}}$%
}

\usepackage{umoline}

\definecolor{mybrown}{RGB}{165, 42, 42}

\definecolor{mypeach}{RGB}{249, 139, 136}

\definecolor{gblue}{RGB}{0, 0, 255}

\definecolor{mygreen}{RGB}{10, 245, 10}

\definecolor{Gray}{gray}{0.9}
\definecolor{green}{rgb}{0.4,0.8,0.25}
\definecolor{red}{rgb}{0.7,0.0,0.0}
\newcommand{\smlt}[2]{\footnotesize{\textcolor{#1}{#2}}}

\begin{document}
\twocolumn[{%
\renewcommand\twocolumn[1][]{#1}%
\title{
    Give Me Your Attention:\\
	Dot-Product Attention Considered Harmful for Adversarial Patch Robustness
}
 \author[1,2]{Giulio Lovisotto}  
 \author[2]{Nicole Finnie}
 \author[2]{Mauricio Munoz}
 \author[2,3]{Chaithanya Kumar Mummadi} 
 \author[2]{Jan Hendrik Metzen}
 \affil[1]{University of Oxford}
 \affil[2]{Bosch Center for Artificial Intelligence}
 \affil[3]{University of Freiburg\authorcr}
 
 \affil[ ]{  \footnotesize \authorcr \tt\footnotesize giulio.lovisotto@cs.ox.ac.uk, \authorcr \tt\footnotesize \{Nicole.Finnie,AndresMauricio.MunozDelgado,ChaithanyaKumar.Mummadi,JanHendrik.Metzen\}@de.bosch.com}

\maketitle
\vspace*{-1.2cm}
\begin{center}
    \centering
    \captionsetup{type=figure}
    \includegraphics[width=.89\linewidth]{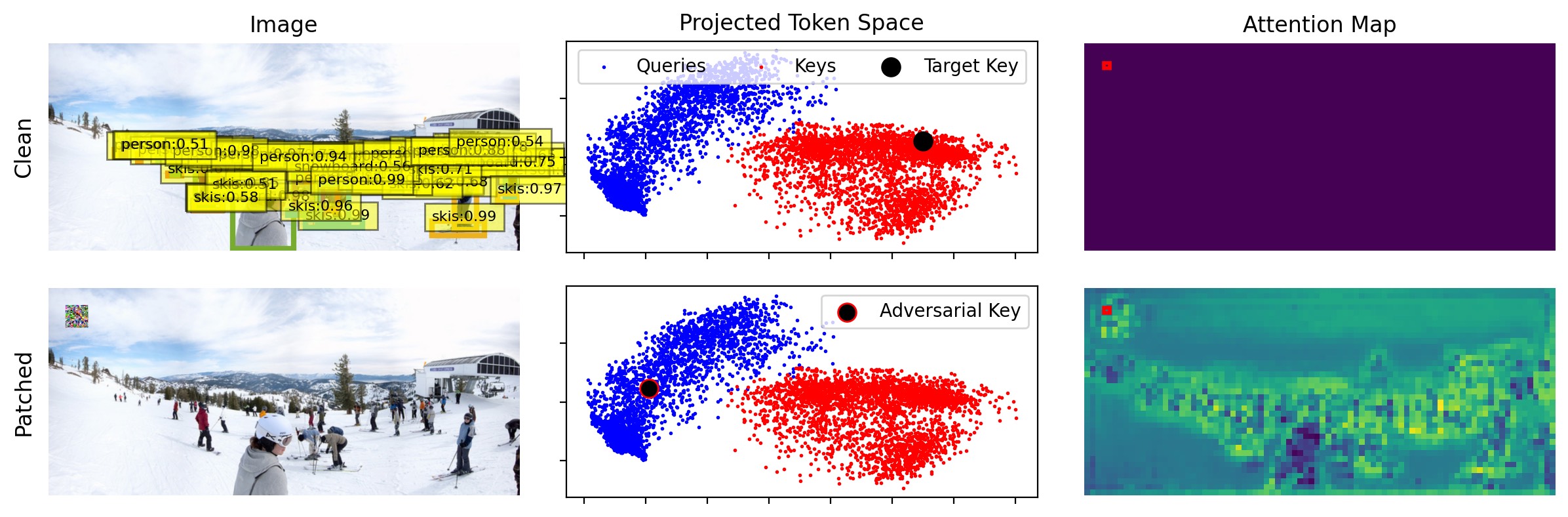}
    \captionof{figure}{Comparison of clean and adversarially patched input for DETR~\cite{carion2020end}. The patch shifts a targeted key token towards the cluster of query tokens (middle column). For dot-product attention, this effectively directs the attention of all queries to the malicious token and prevents the model from detecting the remaining objects. The right column compares queries' attention weights to the target key, marked by a red box, between clean and patched inputs and highlights large attention weights drawn by this adversarial key.}
	\label{fig:teaser}
\end{center}%
}]

\footnotetext{This paper will appear in the proceedings of CVPR22.}

\begin{abstract}
\vspace*{-.5cm}
Neural architectures based on attention such as vision transformers are revolutionizing image recognition. Their main benefit is that attention allows reasoning about all parts of a scene jointly. In this paper, we show how the global reasoning of (scaled) dot-product attention can be the source of a major vulnerability when confronted with adversarial patch attacks. We provide a theoretical understanding of this vulnerability and relate it to an adversary's ability to misdirect the attention of all queries to a single key token under the control of the adversarial patch. We propose novel adversarial objectives for crafting adversarial patches which target this vulnerability explicitly. We show the effectiveness of the proposed patch attacks on popular image classification (ViTs and DeiTs) and object detection models (DETR). We find that adversarial patches occupying 0.5\% of the input can lead to robust accuracies as low as 0\% for ViT on ImageNet, and reduce the mAP of DETR on MS COCO to less than 3\%.

\end{abstract}
\vspace{-.5cm}


\section{Introduction}
\label{sec:intro}
The attention mechanism plays a prominent role in recent success of transformers for different language and image processing tasks. Recent breakthroughs in image recognition using vision transformers \cite{dosovitskiy2020image, pmlr-v139-touvron21a, liu2021swin} have inspired different architectures' design for tackling tasks such as object detection \cite{carion2020end, beal2020transformerbased}, semantic segmentation \cite{strudel2021segmenter, zheng2021rethinking, wang2021end, xie2021segformer}, image synthesis \cite{wang2020sceneformer, esser2021taming, jiang2021transgan}, video understanding \cite{sun2019videobert, girdhar2019video, lee2020parameter, zhou2018end, neimark2021video, girdhar2019video, arnab2021vivit, liu2021video, bertasius2021space}, and low-level vision tasks \cite{yang2020learning, chen2021pre, liang2021swinir, wang2021uformer}. These different transformers use the dot-product attention mechanism as an integral component of their architectural design to model global interactions among different input or feature patches. Understanding robustness of these dot-product attention-based networks against adversarial attacks targeting security-critical vulnerabilities is important for their deployment into real-world applications.

The increasing interest of research in transformers across vision tasks has motivated several recent works \cite{naseer2021intriguing, mahmood2021robustness, aldahdooh2021reveal, naseer2021improving, benz2021adversarial, wei2021towards, bhojanapalli2021understanding, shao2021adversarial, anonymous2022patchfool, anonymous2022patchwise} to study their robustness against 
adversarial attacks. Some prior works \cite{naseer2021intriguing, aldahdooh2021reveal, benz2021adversarial, shao2021adversarial} have hypothesized that transformers are more robust than convolutional neural networks (CNNs) against 
these attacks.
On the other hand, \cite{mahmood2021robustness, wei2021towards, anonymous2022patchfool, anonymous2022patchwise} have shown that vision transformers are not an exception and are also prone to adversarial attacks. In particular, \cite{wei2021towards} shows that an adversarial attack can be tailored to transformers
to achieve high adversarial transferability. These findings indicate that robustness evaluation protocols (attacks) designed for CNNs might be suboptimal for transformers. In the same line of work, we identify a principled vulnerability in the widely-used dot-product attention in transformers  that can often be exploited by image-based adversarial patch attacks.

Dot-product attention computes the dot-product similarity of a query token with all key tokens, which is later normalized using the softmax operator to obtain per token attention weights.
These attention weights are then multiplied with value tokens to control the value token's contribution in the attention block. Gradient-based adversarial attacks backpropagate gradients through all the components in the architecture including the attention weights.
We observe that on pretrained vision transformers, because of the softmax, the gradient flow through the attention weights is often much smaller than the flow through the value token (refer to Sec.~\ref{sec:gradient_of_dot_product_self_attention}). Consequently, gradient-based attacks with a standard adversarial objective are biased towards focusing on adversarial effects propagated through the value token and introduce little or no adversarial effect on the attention weights, thus limiting the attack's potential.

Our work aims to adversarially affect the attention weights, even if those operate in a saturated regime of softmax where gradient-based adversarial attacks are impaired. 
Specifically, we propose losses that support the adversary in misdirecting the attention of most queries to a key token that corresponds to the adversarial patch, i.e., to increase the attention weights of the queries to the targeted key token.
We further study necessary conditions (refer to Sec.~\ref{sec:robustness_of_dot_product}) for a successful attack that misdirects attention weights: both (a) having projection matrices with large singular values and (b) having higher embedding dimensions, allows amplifying the effect of perturbations in a single token, and thus changing the embedding of a single key  considerably.
Moreover, (c) having less centered inputs (larger absolute value of input mean) to the dot-product attention results in distinct clusters of queries and keys that are distant from each other.
Under this condition, moving a single key closer to the query cluster center can make the key most similar to the majority of queries simultaneously (see Figure \ref{fig:teaser}).

\begin{figure*}[t]
	\centering
	\includegraphics[width=0.95\linewidth]{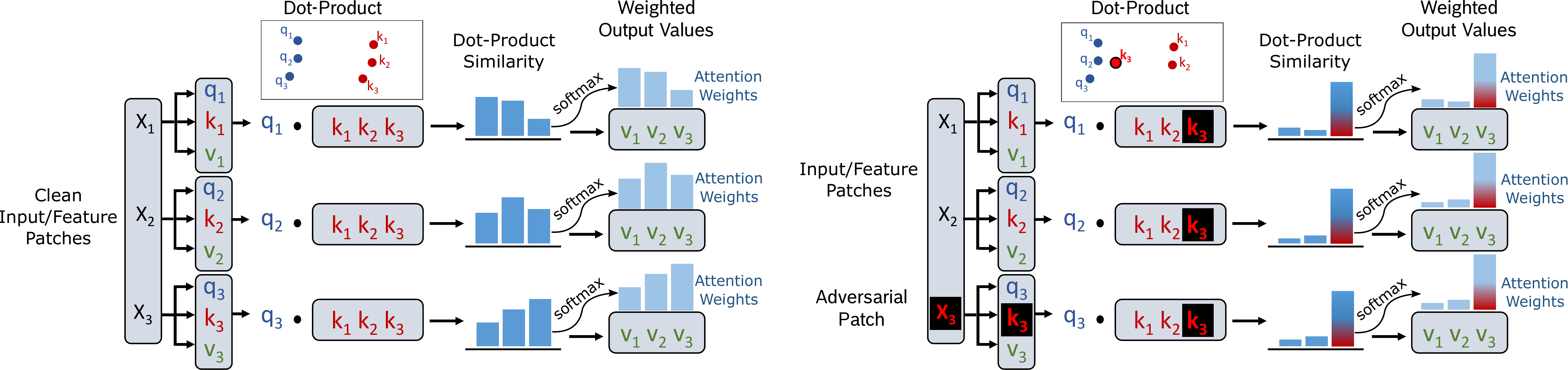}
	\caption{Example of dot-product (self-)attention mechanism for clean (left) and adversarial patch attack (right) settings. Here, $q$, $k$, and $v$ stand for projected queries, keys, and value tokens of input features. Left: dot-product attention computes dot-product similarities of a query with all keys, which is later normalized using softmax to obtain per token attention weights. These are multiplied with value tokens to control their contributions in an attention block. Right: \emph{Attention-Fool} losses optimize the adversarial patch in input at $X_3$ to maximize dot-product similarity of all the queries to the key $k_3$ (marked in red/black) , 
	which corresponds to moving $k_3$ closer to the queries cluster.
	The increase in dot-product similarity of queries with $k_3$ 
	misdirects the model's attention from image content to adversarial patch. 
	} 
	\label{fig:method}
\end{figure*}

We propose a family of adversarial losses and attacks called \emph{Attention-Fool} that directly acts on the dot-product outputs (pre-softmax dot-product similarities).
These losses optimize the adversarial patch to maximize the dot-product similarity of all the queries to a desired key (typically the key whose token corresponds to the input region with adversarial patch).
This approach maximizes the number of queries that attend to the targeted key in an attention head, please refer to Figure~\ref{fig:method} for an illustration.
We apply these losses at multiple attention heads and layers in transformers to misdirect the model's attention from the image content to the adversarial patch in order to encourage wrong predictions.
We show that our \emph{Attention-Fool} adversarial losses improve gradient-based attacks against different vision transformers ViTs~\cite{dosovitskiy2020image} and DeiTs~\cite{pmlr-v139-touvron21a} for image classification and also significantly improve the attack's effective on the object detection model DETR~\cite{carion2020end}.


Our contributions can be summarized as follows: We
\begin{itemize}[itemsep=1pt,parsep=1pt,topsep=3pt]
\item Identify the necessary conditions for the existence of a vulnerability in dot-product attention layers weights, see Section~\ref{sec:robustness_of_dot_product}.
\item Provide reasons why this vulnerability may not be fully exploited by vanilla gradient-based adversarial attacks, see Section~\ref{sec:gradient_of_dot_product_self_attention}.
\item Introduce Attention-Fool, a new family of losses which defines losses directly on the attention layers dot-product similarities, see Section~\ref{sec:attention-fool}. 
\item Show that transformers for image classification and object detection are highly sensitive to small adversarial patches, see Section~\ref{sec:experiments}.
\end{itemize}

\section{Related Work}\label{sec:related_work}
Recent successes of Vision Transformers (ViTs) \cite{dosovitskiy2020image}
have inspired several works \cite{naseer2021intriguing, mahmood2021robustness, aldahdooh2021reveal, naseer2021improving, benz2021adversarial, wei2021towards, shao2021adversarial, anonymous2022patchfool, anonymous2022patchwise} that study their robustness against adversarial attacks. While some works \cite{aldahdooh2021reveal, benz2021adversarial, shao2021adversarial, naseer2021intriguing, naseer2021intriguing} have hypothesized that ViTs are more robust than CNNs under different white- and black-box transfer attack settings (including universal adversarial patches), others \cite{mahmood2021robustness, naseer2021improving, wei2021towards, bhojanapalli2021understanding} have claimed that ViTs are at least as vulnerable as CNNs. 
In particular, \cite{aldahdooh2021reveal} attributed ViT's presumed robustness to its ability to capture global features.
\cite{benz2021adversarial, shao2021adversarial} analyzed that ViTs 
rely on low-frequency features that are robust to adversarial perturbations. \cite{shao2021adversarial} have stated that ViTs have better certified robustness than CNNs 
and have further suggested that adversarially trained ViTs 
have comparable robustness to their CNN counterparts. However, they observed catastrophic overfitting for ViTs when using fast adversarial training \cite{wong2020fast}, suggesting the need for improvements in adversarial training. 

On the other hand, \cite{bhojanapalli2021understanding} have claimed that ViTs pretrained on larger datasets are at least as vulnerable as CNN counterparts. \cite{mahmood2021robustness, naseer2021improving, wei2021towards} suggested that adversarial transferability between ViTs and from ViTs to CNNs can be improved by carefully tailoring adversarial attacks to the transformer architecture. \cite{mahmood2021robustness} explored ensembling of CNNs and ViT models to improve attack transferability. The authors of \cite{naseer2021improving} proposed a self-ensemble technique: split a single ViT model into an ensemble of networks to improve the transferability. \cite{wei2021towards} proposed an attack that skips attention gradients to generate highly transferable adversarial perturbations.

In this work, we aim to understand robustness of the widely-used dot-product attention in transformers and expose its vulnerability to adversarial patch attacks. These are constructed by tailoring adversarial objectives to specifically fool the dot-product attention mechanism.
Concurrent to our work, \cite{anonymous2022patchfool, anonymous2022patchwise} also fool the attention mechanism in transformers using image patches. \cite{anonymous2022patchwise} uses a conventional adversarial patch attack to mislead the model attention to the perturbed patch to promote wrong predictions. They also show that the adversarial patch generalizes to different positions in the image.
\cite{anonymous2022patchfool} optimizes the adversarial patch to increase its attention weights (post-softmax) from all other patches to attack the model.
We discuss limitations of optimizing post-softmax attention weights as in \cite{anonymous2022patchfool} (refer to Section \ref{sec:gradient_of_dot_product_self_attention}) and propose to optimize directly the pre-softmax dot-product similarities of query and keys to draw the attention  to the adversarial patch.

Besides optimizing the content of the patch, there exist also methods for optimizing the location of the patch in the input. Joshi et al.~\cite{joshi2021adversarial} select  the location of adversarial patches based on token saliency. Fu et al.~\cite{anonymous2022patchfool} selects a salient image patch to contain the adversarial patch, based on how much attention the image patch draws in the clean image. Our work abstracts from selecting the patch location and focuses on the loss for the optimization of the patch's content. Any patch location-selection method is orthogonal and could be combined with our Attention Fool-losses.

This work focuses on dot-product attention, widely used in transformers. We leave investigation of vulnerabilities of other attention mechanisms \cite{lee2021fnet, zhu2020deformable, wang2020linformer} to future work.


\section{Preliminaries}\label{sec:limitations-of-std-evaluation}
We introduce the objective for patch robustness evaluation, summarize an optimization algorithm for finding adversarial patches, and recap scaled dot-product attention.

\paragraph{Generic Objective Formulation.}
Given a (normalized) image $x \in [0, 1]^{3\times h \times w}$ and associated label $y$, we craft an adversarial patch $p \in [0, 1]^{3\times p_h \times p_w}$ with $p_h\ll h, p_w\ll w$ that maximizes the following objective:
\begin{equation}\label{eq:formulation}
    \text{arg }\text{max}_{p} \:\: \mathcal{L}(f(\mathcal{F}(x, p, \text{L})), y),
\end{equation}
with $\text{L}$ specifying the location of the patch $p$ within the larger image $x$, $\mathcal{F}$ a function to apply the patch onto the image (i.e., overwriting an input region for a given size), and $f$ being the target model.
For classification tasks, we are interested in the 0-1 loss $\mathcal{L}(x, y) = \mathcal{L}_{0,1}(x, y) = \begin{cases} 0 & x = y \\ 1 & x \neq y \end{cases}$, which corresponds to finding patches that maximize misclassifications.
We note that the constraint $p \in [0, 1]^{3\times p_h \times p_w}$ can be rewritten as $\vert\vert p - 0.5 \vert\vert_\infty \leq 0.5$.

\paragraph{Threat Model.}
We focus on a white-box threat model, where an adversary has access to the model's internals (this includes intermediate network layers outputs).
As in~\cite{anonymous2022patchfool, croce2020sparse}, we do not consider the imperceptibility of the patch to be a requirement.
We also focus on a single-patch fixed-location threat model (L in Equation~\ref{eq:formulation} is fixed a priori). 
Note that methods for choosing the patch location~\cite{anonymous2022patchfool, joshi2021adversarial} could be combined with the proposed approach.

\paragraph{Optimization Algorithm.}
For the patch optimization, throughout the paper, we use Projected Gradient Descent (PGD)~\cite{madry2017towards} for $\ell_\infty$-norm bounded perturbations:
\begin{equation}\label{eq:pgd}
    p^{t+1} = p^{t} + \alpha \cdot \text{sgn}\;(\nabla_p \mathcal{L}(f(\mathcal{F}(x, p, \text{L}), y))).
\end{equation}
We initialize $p^0$ uniform randomly from $ [0, 1]^{3\times p_h \times p_w}$.
Since the 0-1 loss $\mathcal{L}_{0,1}$ is piecewise-constant, it is not suited for gradient-based optimization. Accordingly, a surrogate loss such as the cross-entropy $\mathcal{L}$=$\mathcal{L}_{\text{ce}}$ is used typically. However, we propose alternative losses in Section \ref{sec:attention-fool}.

\paragraph{Dot-Product Attention.} 
In its basic form, dot-product attention~\cite{DBLP:journals/corr/LuongPM15, vaswani2017attention} computes, for every query, attention weights as the dot-product of the query to all keys. The softmax function is then applied over the key dimension. These attention weights are then multiplied by the values:
\begin{equation}
	\text{Attention}(Q, K, V) = \text{softmax}(QK^\top)V.
\end{equation}
Here, $Q \in \mathbb{R}^{n \times d_{\text{model}}}$, $K \in \mathbb{R}^{n \times d_{\text{model}}}$, and $V \in \mathbb{R}^{n \times d_{\text{model}}}$ are the matrices of $n$ queries, keys, and values, respectively. According to Vaswani et al.~\cite{vaswani2017attention}, for large values of $d_{\text{model}}$, the dot-product between queries and keys may grow large in magnitude. This has the effect of pushing the softmax function into the saturated regime, where it has extremely small gradients. This is due to the exponentiation of individual query-key dot-products in the softmax function.
Because this can be harmful for training, they introduce scaled dot-product attention, where $QK^\top$ is scaled by $\nicefrac{1}{\sqrt{d_{\text{model}}}}$.

In practice, using $H>1$ attention heads by linearly projecting queries, keys, and values $H$ times to $d_k$, $d_k$, and $d_v$ dimensions was found to be beneficial~\cite{vaswani2017attention}.
The output of the $h$-th attention head (AH) becomes:
\begin{equation}
	\text{AH}_h(Q,K,V) = \text{softmax}(\frac{QW^h_Q(KW^h_K)^\top}{\sqrt{d_k}}) VW^h_V,
\end{equation}
where $W^h_Q \in \mathbb{R}^{d_{\text{model}}\times d_k}$ , $W^h_K \in \mathbb{R}^{d_{\text{model}}\times d_k}$, $W^h_V \in \mathbb{R}^{d_{\text{model}}\times d_v}$ are (learned) projection matrices.
The outputs of individual attention heads are concatenated and multiplied by another learned projection matrix $W_O \in \mathbb{R}^{Hd_v\times d_\text{model}}$.

A special case is self-attention with $Q=K=V\in \mathbb{R}^{n \times d_\text{model}}$, which is typically used in encoder layers of image recognition models. We define the attention weights of the $h$-th head on $X$ via $A_h(X) = \text{softmax}(\frac{XW^h_Q(XW^h_K)^\top}{\sqrt{d_k}}) \in \mathbb{R}^{n\times n}$. The $h$-th self-attention head becomes:
\begin{equation}
	\text{SelfAH}_h(X) = A_h(X) XW^h_V
\end{equation}

\section{Robustness of Dot-Product Attention}\label{sec:method}
In this section, we first discuss why (scaled) dot-product self-attention is challenging for gradient-based adversarial attacks such as \pgd. We then provide an example of an adversarial vulnerability in the attention weights themselves.

\subsection{Gradient of Dot-Product Self-Attention} \label{sec:gradient_of_dot_product_self_attention}
For computing $\nabla_p \mathcal{L}(\mathcal{F}(x, p, \text{L}), y)$ in Eq.~\ref{eq:pgd}, it is necessary to backpropagate through the entire model. This requires the gradient $\nabla_X \text{SelfAH}_h(X)$ for every attention layer and head $h$. With the product-rule we obtain:
$$\nabla_X \text{SelfAH}_h(X) = \left[(\nabla_X A_h(X))  X + A_h(X) 1_X \right] W^h_V,$$
where $1_X$ is a matrix of ones with the same shape as X.

\begin{table}[bt]
	\centering
	
	\caption{Median of $| (\nabla_X A_h(X))  X) / (A_h(X)  1_X) |$ over tokens and heads on a random natural image for models on 6 encoder layers. A larger version of this table is in Section~\ref{tab:grad_ratio_full} in Appendix.}
	\footnotesize
	\label{tab:gradient_ratio}
\begin{tabular}{lcccccc}\toprule
        & ViT-T & ViT-S & ViT-B & DeiT-T & DeiT-S & DETR
        \\\midrule
          Layer 1 & 0.048 & 0.059 & 0.064 & 0.044 & 0.049 & 0.188 \\
          Layer 2 & 0.032 & 0.027 & 0.032 & 0.042 & 0.033 & 0.040 \\
          Layer 3 & 0.035 & 0.032 & 0.028 & 0.034 & 0.028 & 0.034 \\
          Layer 4 & 0.029 & 0.035 & 0.036 & 0.035 & 0.029 & 0.058 \\
          Layer 5 & 0.041 & 0.032 & 0.048 & 0.066 & 0.036 & 0.074 \\
          Layer 6 & 0.030 & 0.029 & 0.036 & 0.040 & 0.034 & 0.112 \\
        \bottomrule
    \end{tabular}
\end{table}

An important property of the gradient  $\nabla_X \text{SelfAH}_h(X)$  is accordingly the element-wise ratio $| \frac{(\nabla_X A_h(X)) X}{A_h(X) 1_X}|$.
We summarize the median value of this ratio over tokens and heads in Table \ref{tab:gradient_ratio} for different models and layers. As seen, the typical regime (that is, for $> 50\%$ of the cases) is that $(\nabla_X A_h(X))  X$ is smaller than $A_h(X)1_X$ by a factor of $\approx 20$   for  ViTs \cite{dosovitskiy2020image}, DeiTs \cite{pmlr-v139-touvron21a}, and for the inner encoder layers of DETR \cite{carion2020end}. In this setting, one can approximate $\nabla_X \text{SelfAH}_h(X) \approx (A_h(X)  1_X) W^h_V$, that is: the gradient considers the attention weights $A_h(X)$ as effectively constant. Accordingly, gradient-based attacks such as \pgd{} based on an end-to-end loss such as $\mathcal{L}_{ce}$ would be biased towards focusing on adversarial effects in $X$ that can be propagated (linearly) via the values $V=XW^h_V$ in self-attention, while effectively ignoring potentially adverse (and non-linear) effects of $X$ propagated via the attention weights $A_h(X)$. We note that also later stages of model training with gradient-based optimizers can be negatively affected by this property of dot-product attention. Studying this in more detail is left to future work.

\begin{figure*}[t]
	\centering
	\includegraphics{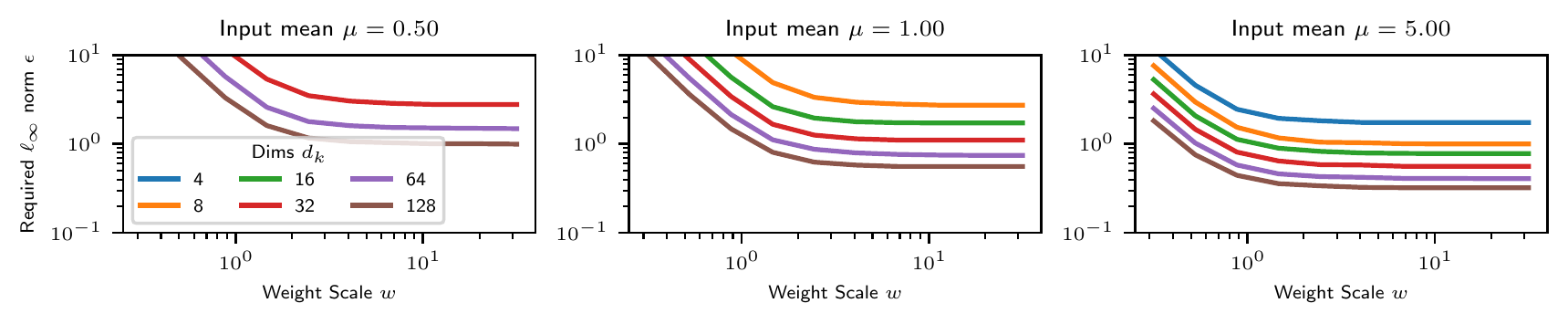}
	\vspace*{-.75cm}
	\caption{Minimum $\ell_\infty$ norm perturbation $\epsilon$ required for reaching the attacker's goal (i.e., forcing queries to attend to the first key) on the controlled setting from Section \ref{sec:robustness_of_dot_product}. Increased weight scale $w$, higher embedding dimension $d_k$, and larger input mean norm $\mu$ simplify the attack: as the three quantities increase, the $\epsilon$ perturbation required to fulfil the goal decreases.
	} 
	\label{fig:toy_example}
\end{figure*}

\subsection{Robustness of Dot-Product Attention Weights}
\label{sec:robustness_of_dot_product}
We now study to which extent attention weights $A_h(X)$ of dot-product attention can be affected by an adversarial patch attack. For this, we use a controlled setting with normally distributed $X$, where each feature has mean $\mu$ and variance $1$: $X_j \sim \mathcal{N}(\mu \cdot \mathbf{1}, \mathbf{1})$. Moreover, we choose $d_k = d_\text{model}$ and diagonal $W_Q = -w \cdot \mathbb{I}_{d_k}$ and $W_K = w \cdot \mathbb{I}_{d_k}$, that is $W_Q$ and $W_K$ having both scale $w$ but opposite signs. 

We study a simple threat model: the adversary can only modify $X_0$, the first of $n$ entries of $X$ ($X$ can be considered as embedding of patches and $X_0$ as corresponding to the embedded adversarial patch), with the constraint $\vert\vert X_0^{adv} - X_0 \vert\vert_\infty \leq \epsilon$. Moreover, the adversaries goal is to achieve $\left[ \frac{1}{n} \sum_j \left(A_h(X^{adv})_{0j} \geq 0.99 \right)\right] > 0.95$, that is: at least $95\%$ of the queries need to attend to the first key with attention weights greater or equal $0.99$. By design, setting $X_0^{adv} = X_0 - \epsilon$ is a strong attack for $\mu > 0$ because $-\mathbf{1} \cdot \epsilon$ corresponds to the direction of $(W_Q  - W_k)(\mu \cdot \mathbf{1})$, the difference of projected query and key mean.

We study how $\epsilon$ needs to be chosen as a function of $\mu$, $w$, and $d_k$ for a successful attack with the above attack and threat model. Results are shown in Figure \ref{fig:toy_example}. In general, one can observe that a successful attack with a smaller perturbation amount $\epsilon$ requires: (a) increasing the scale $w$ of the projection matrices, (b) higher embedding dimensions $d_k$, and (c) less centered inputs X (larger $|\mu|$). The findings (a) and (b) can be attributed to higher dimensions and larger weights allow the effect of minor perturbations in each input dimension to be amplified.
Upon closer inspection (see Section~\ref{app:silhouette_score}
in Appendix),
we can attribute finding (c) to a separation of projected keys and queries into distinct clusters for less centered inputs: all queries are close to each other, all keys are close to each other, but query-key pairs are all distant from each other. In this case, a single key can be made to be the most similar to each query, just by moving this key in the direction of the query cluster (see Figure ~\ref{fig:teaser} for a real data illustration) .

\begin{table}
	\centering
	
	\caption{Largest singular value of the projection weight matrices $W_Q(W_K)^\top$ for randomly initialized and trained models.}
	\footnotesize
	\label{tab:singular_values}
\begin{tabular}{l|ccc|cc}\toprule
         & ViT/DeiT-B & ViT-B & DeiT-B & \multicolumn{2}{c}{DETR encoder} \\
         & random   & trained & trained & random &  trained \\
         \midrule
          Layer 1 & 0.80 & 60.96 & 175.28 & 1.27 & 7.93 \\
          Layer 2 & 0.79 & 22.67 & 65.86 & 1.29 & 10.61 \\ 
          Layer 3 & 0.80 & 11.76 & 54.19 & 1.25 &  15.35  \\
          Layer 4 & 0.79 & 5.88 & 44.95 & 1.27 & 45.26  \\
          Layer 5 & 0.79 & 4.91 & 29.36 & 1.28 & 49.82 \\
          Layer 6 & 1.21 & 4.83 & 28.72 & 1.26 & 29.30 \\
        \bottomrule
    \end{tabular}
\end{table}

In which regime does dot-product attention typically operate in trained image transformers? For many architectures, $d_k$ is relatively large by design. Moreover, the input of dot-product attention is typically not centered (zero-mean) due to enabled affine transformations within the normalization layers before self-attention. Many implementations, e.g.~\cite{pytorch-multiheadattention}, also use affine rather than linear query/key projections in the heads, where different biases in the projection of key and queries will have a similar effect as non-centered inputs.
Lastly, we also observe that for many vision transformers, the product of projection weight matrices $W_Q(W_K)^\top$ has very large maximal singular value at convergence, compared to the randomly initialized projection weight matrices (see Table \ref{tab:singular_values}). Large singular values can be considered to be analogous to the ``large weight scales'' in our controlled setting.
We thus expect that dot-product attention of trained vision transformers typically operates in a setting where the attention weights $A_h(X)$ can be a source of vulnerability to patch attacks, but gradient-based attacks such as \pgd{}  are biased towards ignoring this vulnerability (as discussed in Sec. \ref{sec:gradient_of_dot_product_self_attention}).



\section{Attention-Fool}\label{sec:attention-fool}

\begin{figure*}[t]
	\centering
	
	\includegraphics[width=0.99\linewidth]{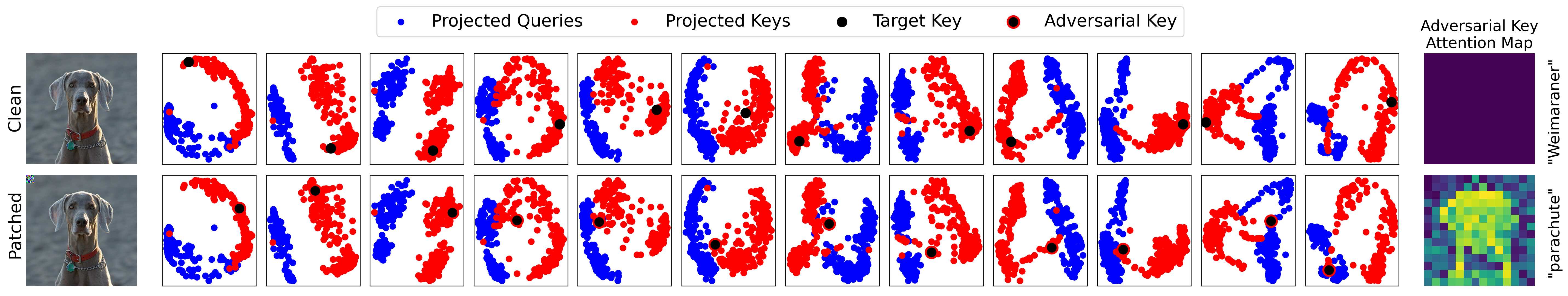}

	\caption{Embedded projected key and query tokens for clean and patched input images on each of 12 ViT layers of the attention head which showed the largest attention change. We design Attention-Fool to target all layers and heads at once. The last column reports the attention map weights of the adversarial key on the last layer -- showing that the key draws a large amount of attention from queries.  }
	\label{fig:vit-per-layer-visualization}
\end{figure*}

We have discussed that attention weights $A_h(X)$ can be largely affected by adversarial patches in principle but we have also shown that gradient-based attacks are impaired in exploiting this vulnerability.
In order to encourage fooling of attention weights, we introduce a family of losses that are defined directly on the pre-softmax attention logits $B_h(X)=\frac{XW^h_Q(XW^h_K)^\top}{\sqrt{d_k}}$. 
We denote attacks based on these losses as \emph{Attention-Fool}.
We note that Attention-Fool losses are always maximized by the attacker, regardless of whether an attack is targeted or untargeted.
Moreover, while this paper focuses on self-attention, it is straightforward to extend these losses to cross-attention.

\paragraph{$\mathcal{L}^{hl}_{kq}$: Attacking a single attention head.} 
We first focus on adversarially modifying attention weights of a specific head $h$ in a specific self-attention layer $l$. We propose a loss $\mathcal{L}^{hl}_{kq}$ that aims at maximizing the query attention to the $i^\star$-th projected key, that is: maximize the number of queries that devote the majority of their attention to this key.  Naturally, $i^\star$ is chosen such that its token corresponds to the region in the input where the adversarial patch has been placed.

We denote projected queries by $P_Q^{hl} = X^{hl}W_Q^{hl} \in \mathbb{R}^{n \times d_k}$ and projected keys by $P_K^{hl} = X^{hl}W_K^{hl} \in \mathbb{R}^{n \times d_k}$. We have $B^{hl} = \frac{P_Q^{hl} (P_K^{hl})^\top}{\sqrt{d_k}} \in \mathbb{R}^{n \times n}$, where the first dimension indexes queries and the second one keys. Each element of $B^{hl}$ quantifies the dot-product similarity between a key and a query in the respective attention head $h$ and layer $l$.
We  now set $\mathcal{L}^{hl}_{kq} = \frac{1}{n}\sum_{j} B^{hl}_{ji^\star}$, where $j$ indexes the queries. Maximizing $\mathcal{L}^{hl}_{kq}$ thus corresponds to maximizing average dot-product similarity between queries and the target key.


\paragraph{$\mathcal{L}_{kq}$: Attacking all attention layers and heads simultaneously.}
In the previous paragraph, we have introduced a loss that targets a single head and layer.
However, in general fooling a single head might not be sufficient and it is also a priori unclear which head would be the most vulnerable one.
The same applies to the choice of the layer, when there are multiple layers using dot-product attention. Because of this, we propose applying the loss to all layers and heads simultaneously.
However, simply averaging $\mathcal{L}^{hl}_{kq}$ over all heads and layers might not be optimal because it may favour many smaller head-wise changes overall rather than successfully fooling a subset of heads and layers to a larger extent.
Because of this, we will utilize a smooth maximum $\smax(x) = \log \sum_i e^{x_i}$ over $\mathcal{L}^{hl}$. Specifically, we define 
$\mathcal{L}^l_{kq} = \log \sum_h e^{\mathcal{L}^{hl}_{kq}} \quad \forall l$ and $\mathcal{L}_{kq} = \log \sum_l e^{\mathcal{L}^{l}_{kq}}$. We empirically compare the choice of this smooth maximum over mean and hard maximum in Section \ref{sec:vit_ablation} and the choice of $l$ in $\mathcal{L}^l_{kq}$ in Section~\ref{app:lkq-vit-layer-and-head}. 
We visualize $\mathcal{L}_{kq}$'s effect on each network layer in Figure~\ref{fig:vit-per-layer-visualization}.

In order to make the losses of different heads and layers commensurable, we also propose to normalize scales of projected keys and queries via $\bar{P}^{hl}_Q = \nicefrac{P^{hl}_Q}{\frac{1}{n} \vert\vert P^{hl}_Q\vert\vert_{1, 2}}$ and $\bar{P}^{hl}_K = \nicefrac{P^{hl}_K}{\frac{1}{n} \vert\vert P^{hl}_K\vert\vert_{1, 2}}$, where $\vert\vert X \vert\vert_{1,2} = \sum\limits_{i}^n\sqrt{\sum\limits^{d_k}_{j} X_{ij}^2}$ is the $L_{1, 2}$ norm.
Note that the normalization is applied per individual head and makes the average $\ell_2$ norm of queries and keys equal to 1. $B^{hl}$ is then computed based on $\bar{P}^{hl}_Q$ and $\bar{P}^{hl}_K$, making the losses $\mathcal{L}^{hl}_{kq}$ commensurable.
We empirically evaluate the effect of this normalization in Section~\ref{app:sec:normalization-vit} of the supplementary material.

\paragraph{$\mathcal{L}_{kq^\star}$: Targeting a special class token.}
The $\mathcal{L}_{kq}$ loss treats all queries equally and aims at misdirecting the attention of the majority of queries towards the adversarial key token.
However, for many architectures not all queries are created equally, for instance for ViTs \cite{dosovitskiy2020image} and DeiTs \cite{pmlr-v139-touvron21a}, there exists a special class token, which is supposed to accumulate class evidence over the layers.
We propose a version of the $\mathcal{L}_{kq}$ loss that specifically targets the attention of a certain query, for instance the one corresponding to the class token.
Let $j^\star$ be the index of the query that shall be targeted.
We define $\mathcal{L}^{hl}_{kq^\star} =  B^{hl}_{j^\star i^\star}$ and generalize to $\mathcal{L}_{kq^\star}$ as above using smooth maximum over heads and layers.

In the following experiments, we will use combinations of $\mathcal{L}_{kq}$ or $\mathcal{L}_{kq^\star}$  with standard cross-entropy loss $\mathcal{L}_{ce}$ for the adversarial patch optimization of Equation~\ref{eq:formulation}. We denote the resulting set of attacks leveraging the weakness in dot-product attention as \emph{Attention-Fool}. Note that Attention-Fool is different from concurrent work Patch-Fool~\cite{anonymous2022patchfool} that defines a loss on the post-softmax attention weights, averaged of heads $h$ and queries $j$: $\mathcal{L}_{PF} = \sum_h \sum_j A_h(X)_{j}$. In particular, Attention-Fool (in contrast to Patch-Fool) is not affected by the small gradient issue described in Section \ref{sec:gradient_of_dot_product_self_attention} since it is defined on pre-softmax attention weights.

\begin{table*}[t]
        \centering
        \caption{Robust accuracies (\%) under adversarial patch attacks on 1000 ImageNet images. Clean performance, as well robust accuracy against Patch-Fool loss~\cite{anonymous2022patchfool} and Patch-RS~\cite{croce2020sparse} are shown as baselines. All rows in the bottom block are computed using \pgd$^{250}$ with step size $\alpha$=8/255 and crossentropy loss $\mathcal{L}_{ce}$. An Attention-Fool term is added to the loss (either $\mathcal{L}_{kq}$ or $\mathcal{L}_{kq\star}$) and optionally momentum is added to \pgd. Numbers in parenthesis report the improvement or degradation in robust accuracy w.r.t. the $\mathcal{L}_{ce}$ baselines. All models use  224$\times$224 resolution unless marked by ``384'' in the model name, then they use 384$\times$384. Patch size is always 16$\times$16.}
        \small 
        \label{tab: untargeted-main}

    \begin{tabular}{lccccccc}\toprule
            & ResNet50 & ViT-T & ViT-B & ViT-B-384 & DeiT-T & DeiT-B & DeiT-B-384
            \\\midrule
        \rowcolor{Gray} Clean & 80.6 & 73.5 & 85.0 & 86.4 & 69.4 & 82.0 & 82.0 \\
        Patch-Fool loss~\cite{anonymous2022patchfool} 
                 & - &  0.3 & 17.5 & 47.8 & 3.2  &  52.9 & 68.2 \\ 
        \rowcolor{Gray} Patch-RS~\cite{croce2020sparse} & 70.8 & 18.6 & 49.8 & 65.9  & 44.0 & 57.3 & 71.2
            \\\midrule
        \pgd$^{250}$ with $\mathcal{L}_{ce}$ & 55.1 & 0.1 & 13.5 & 31.2 & 19.8 & 36.0 & 58.8\\
        \rowcolor{Gray}$\quad + \mathcal{L}_{kq}$ & - & 0.5 \smlt{red}{(+0.4)} & 5.0 \smlt{green}{(--8.5)} & 18.0 \smlt{green}{(--13.2)} & 13.1 \smlt{green}{(--6.7)} & 35.5 \smlt{green}{(--0.5)} & 55.1 \smlt{green}{(--3.7)}\\
        $\quad + \mathcal{L}_{kq\star}$ & - & 0.3 \smlt{red}{(+0.2)} & 2.6 \smlt{green}{(--10.9)} & 13.0 \smlt{green}{(--18.2)} & 11.7 \smlt{green}{(--8.1)} & 33.7 \smlt{green}{(--2.3)} & 57.2 \smlt{green}{(--1.6)}\\
        \rowcolor{Gray}$+$ Momentum & 49.0 & 0.0 & 3.1 & 13.2 & 1.5 & 16.8 & 41.7\\
        $\quad + \mathcal{L}_{kq}$ & - & 0.0 \smlt{green}{(-0.0)} & 0.1 \smlt{green}{(--3.0)} & 2.5 \smlt{green}{(--10.7)} & 0.0 \smlt{green}{(--1.5)} & 19.3 \smlt{red}{(+2.5)} & 39.8 \smlt{green}{(--1.9)}\\
        \rowcolor{Gray}$\quad + \mathcal{L}_{kq\star}$ & - & 0.0 \smlt{green}{(-0.0)} & 0.1 \smlt{green}{(--3.0)} & 1.9 \smlt{green}{(--11.3)} & 0.0 \smlt{green}{(--1.5)} & 13.1 \smlt{green}{(--3.7)} & 40.6 \smlt{green}{(--1.1)}\\
        \bottomrule
        \end{tabular}

    \end{table*}

\section{Evaluation of Attention-Fool} \label{sec:experiments}
We evaluate the robustness of different vision transformers against adversarial patches generated using our proposed Attention-Fool adversarial losses.
We first investigate the robustness of ViT~\cite{dosovitskiy2020image} and the improved DeiT~\cite{pmlr-v139-touvron21a} in Section~\ref{sec:exp-vit}.
Then we show how Attention-Fool generalizes to DETR~\cite{carion2020end} in Section~\ref{sec:exp-detr}, which uses a hybrid CNN plus Transformer architecture to perform object detection.

\subsection{Evaluation Setup}
We use \pgd{}  as the optimizer to solve Equation~\ref{eq:formulation}. We set \pgd{}'s initial step size $\alpha^0=8/255$ and schedule it with a cosine decay:
$ \alpha^{(t+1)} = \alpha^{(0)} \frac{1}{2} (1+\cos{(\pi  \frac{t}{\text{N}})}).$
Additionally, we add a stronger baseline to the adversarial patch robustness evaluation, we use an improved version of \pgd{}{} by adding normalized-momentum in the optimization~\cite{Dong_2018_CVPR}:
\begin{equation*}
    m^{(t)} =  \beta m^{(t-1)} + (1-\beta) \nicefrac{\nabla_p^{t}}{||\nabla_p^{t}||_2},
\end{equation*}
and use $m^{(t)}$ instead of the gradient $\nabla_p^{t}$ in \pgd{}. 

\subsection{Attention-Fool on Vision Transformer} \label{sec:exp-vit}

We use the new Attention-Fool attack described in Section~\ref{sec:attention-fool} to compare the effectiveness of adversarial patches optimized with the new losses $\mathcal{L}_{kq}$ and $\mathcal{L}_{kq\star}$ with adversarial patches optimized using cross-entropy loss $\mathcal{L}_{ce}$.
We use 1,000 images from the ImageNet 2012~\cite{ILSVRC15} dataset and perform an untargeted attack, where $y$ in Eq.~\ref{eq:formulation} is the image ground truth. We also report the results on a targeted attack in Section~\ref{app:lkq-vit-targeted} in Appendix.
For the  experiment, we place an adversarial patch of 16$\times$16 pixels in the top left corner, we investigate different location and sizes in Section~\ref{app:lkq-vit-center} and~\ref{app:sec:lkq-patch-sizes}, respectively.
For the evaluation, we use pre-trained ViT models from \texttt{timm}~\cite{timm-library}.
We choose models which split input images into 16$\times$16 image sub-patches; this way the adversarial patch occupies exactly one of the image sub-patches.
We consider models with input image resolution of 224$\times$224 or 384$\times$384; the adversarial patch covers 0.5\% and 0.17\% of the input, respectively.
For comparison, we also perform the attack on a CNN, ResNet50, using the same setup but optimizing adversarial patches with $\mathcal{L}_{ce}$ only.
We report the resulting robust accuracies in Table~\ref{tab: untargeted-main}.
We find that Attention-Fool's $\mathcal{L}_{kq}$ and $\mathcal{L}_{kq\star}$ improve on the baseline cross-entropy on most models regardless of momentum of \pgd.
While Table~\ref{tab: untargeted-main} shows that $\mathcal{L}_{kq}$ already decreases robust accuracies across most settings, using the architecture-specific Attention-Fool loss-variant $\mathcal{L}_{kq\star}$ brings more stable robust accuracy down among all models, often with large performance gains over cross-entropy alone.
Attention-Fool can bring accuracies down to 0\% in ViT-T and DeiT-T models, and as low as 1.90\% even in ViT-B-384, where the patch occupies only 0.17\% of the input image. Notably, in contrast to prior work (see Sec.~\ref{sec:related_work}) we find all ViT/DeiT variants to be considerably \emph{less} robust than a ResNet50 when the patch is placed in the top left corner, whereas placing the patch in the center could have a larger effect on a CNN.

We include additional comparisons with the ``Attention-Aware Loss'' 
of Patch-Fool (Section~4.4 in~\cite{anonymous2022patchfool}).
For a controlled comparison with Patch-Fool, we perform two adjustments: (i) we disregard the saliency-based selection of the patch location (Section~4.3 in~\cite{anonymous2022patchfool}) and (ii) we replace Adam with \textsc{pgd} (momentum = 0.9). 
This abstracts from attack specifics and allows us to compare losses directly.
The results show that Patch-Fool's Attention-Aware loss underperforms the Attention Fool losses; confirming the limitations of optimizing post-softmax attention weights (discussed in Sec.~\ref{sec:gradient_of_dot_product_self_attention}). Moreover, we compare to Patch-RS~\cite{croce2020sparse}, a random-search based black-box patch attack.  The inferior results of Patch-RS indicate that gradient-free attacks are  no viable alternative for exploiting the existent vulnerabilities in attention weights.

\subsection{Attention-Fool on DETR}\label{sec:exp-detr}
Since Attention-Fool targets dot-product attention layers, it can be adapted to target various tasks and architectures that use this attention.
Here, we apply Attention-Fool to object detection with DETR~\cite{carion2020end}, which combines a CNN backbone with a Transformer encoder-decoder. 

\noindent\textbf{Attention-Fool Configuration.}
While $\mathcal{L}_{kq\star}$ showed superior results for ViT/DeiT in Section~\ref{sec:exp-vit}, it is not applicable to DETR due to the absence of a class token in DETR.  In contrast, we use $\mathcal{L}^{(1)}_{kq}$: a loss targeting all queries and heads but only the first transformer encoder layer ($l=1$) rather than all at once. We focus on $l=1$ because we hypothesize that the transition from CNN to transformer encoder in DETR is specifically brittle due to large activations from the CNN backbone. For a similar reason, we aggregate losses across attention heads using the maximum function rather than $\smax$ used in Section~\ref{sec:method}.

\begin{table}
	\centering
	
	\caption{Mean Average Precisions (mAP) in presence of adversarial patches computed with $\mathcal{L}_{ce}$ and patches computed with Attention Fool $\mathcal{L}_{kq}^{(1)}$, on \textsc{detr} models. The Baseline Clean mAP is obtained on the clean set of images. Three different patch sizes and four different \textsc{detr} models are considered, based either on \textsc{r50} or \textsc{r101} backbone and with and without dilation \textsc{dc5}.}
	\footnotesize
	\setlength{\tabcolsep}{1.5pt}
	\label{tab:detr-targeted4}
\begin{tabular}{lcccc}\toprule
        & \textsc{r50} & \textsc{dc5-r50} & \textsc{r101} & \textsc{dc5-r101}
        \\\midrule

    Clean mAP & 53.00 & 54.25 & 54.41 & 56.74 \\ \midrule 
    64$\times$64: $\mathcal{L}_{ce}$ & 34.91 & 17.67 & 31.76 & 34.71 \\ 
    \rowcolor{Gray}$\ \ + \mathcal{L}_{kq}^{(1)}$ & 21.03 \smlt{green}{(-13.88)} & 7.34 \smlt{green}{(-10.33)} & 2.07 \smlt{green}{(-29.70)} & 4.19 \smlt{green}{(-30.52)} \\ 
    $\ \ \mathcal{L}_{kq}^{(1)}$ only & 29.52 \smlt{green}{(-5.39)} & 15.03 \smlt{green}{(-2.64)} & 2.05 \smlt{green}{(-29.71)} & 10.18 \smlt{green}{(-24.53)} \\

    \rowcolor{Gray} 56$\times$56: $\mathcal{L}_{ce}$ & 37.74 & 24.13 & 32.06 & 38.28 \\ 
    $\ \ + \mathcal{L}_{kq}^{(1)}$ & 28.19 \smlt{green}{(-9.55)} & 14.68 \smlt{green}{(-9.45)} & 3.30 \smlt{green}{(-28.76)} & 6.82 \smlt{green}{(-31.46)} \\ 
    \rowcolor{Gray}$\ \ \mathcal{L}_{kq}^{(1)}$ only & 38.14 \smlt{red}{(+0.40)} & 20.18 \smlt{green}{(-3.95)} & 5.60 \smlt{green}{(-26.46)} & 14.27 \smlt{green}{(-24.01)} \\

    48$\times$48: $\mathcal{L}_{ce}$ & 42.08 & 27.69 & 37.89 & 41.32 \\ 
    \rowcolor{Gray}$\ \ + \mathcal{L}_{kq}^{(1)}$ & 34.88 \smlt{green}{(-7.20)} & 18.94 \smlt{green}{(-8.75)} & 9.80 \smlt{green}{(-28.09)} & 17.05 \smlt{green}{(-24.26)} \\ 
    $\ \ \mathcal{L}_{kq}^{(1)}$ only & 42.19 \smlt{red}{(+0.12)} & 32.60 \smlt{red}{(+4.91)} & 12.85 \smlt{green}{(-25.04)} & 17.55 \smlt{green}{(-23.77)} \\

        \bottomrule
    \end{tabular}

\end{table}

	



    



\noindent\textbf{Evaluation Setup.}
We evaluate a targeted Attention-Fool attack on DETR by choosing the target to be the background class, 
which effectively forces missed detections (i.e., false negatives).
We evaluate four pretrained DETR models from the official repository~\cite{detr-github}, with backbone either ResNet50 or ResNet101, and with or without a dilation in the ResNet layer 5 convolution (\textsc{dc5}).
Note that \textsc{dc5} models have twice the resolution in the backbone-extracted feature map, which is used as the encoder input.
We select 100 images from the MS COCO 2017 validation set~\cite{lin2014microsoft}, and use the default DETR validation image loader which re-scales images to have shortest side 800 pixels.
We evaluate Attention-Fool by targeting a single key token -- a single unit in the CNN feature map -- out of the entire token sequence which is at least 2,500- and 625-tokens long for \textsc{dc5} and non-\textsc{dc5} models (with final length depending on the input image resolution).
We provide an ablation in Section~\ref{app:adv-key-replacement} in appendix where we attribute the adversarial effect to this token alone.
We place the adversarial patch in the image top-left centered around the 80,80 pixel.
Accordingly, we target the key token with index 2,2 in non-dilated models and 4,4 in dilated ones -- this ensures that the key token receptive field is centered around the adversarial patch location in the image.
We record the primary COCO challenge metric, the mean Average Precision (mAP), which averages precision-recall curves scores at various Intersection-over-Union (IoU) thresholds.
To not bias the evaluation towards local patch effects, we ignore detection boxes whose intersection with the patch box is $>$50\% of the detection box area.
We do the same for ground truths for a fair comparison between clean and patched inputs.

\noindent\textbf{Results.}
Similarly to the previous section, we use $\mathcal{L}_{ce}$ as an attack baseline and test the improvement (degradation) in mAP as we combine it with the Attention-Fool loss $\mathcal{L}_{kq}^{(1)}$.
We test three different patch sizes, 64$\times$64, 56$\times$56, and 48$\times$48, which occupy $<$0.64\%, $<$0.49\% and $<$0.36\% of the input image, respectively.
Given the higher complexity of DETR (compared to ViT), we increase the number of \pgd{} iterations to 1000 in this experiment.
We report the resulting mAPs in Table~\ref{tab:detr-targeted4}: baseline mAP on clean images,  mAP under a targeted attack which uses $\mathcal{L}_{ce}$, and the mAP change when we use the combined Attention-Fool $\mathcal{L}_{ce}+\mathcal{L}_{kq}^{(1)}$ loss.
Table~\ref{tab:detr-targeted4} shows that the addition of $\mathcal{L}_{kq}^{(1)}$ reduces the mAP performance across all models and all patch sizes.
We find that larger models with \textsc{detr101} are more vulnerable to Attention-Fool, where the mAP can be reduced down to 2.07 -- this corresponds to suppressing the detection of the vast majority of objects. See Figure~\ref{fig:teaser} for an illustration.

Table~\ref{tab:detr-targeted4} also presents a setting where we do not use any loss on the model's output but solely focus on misdirecting the first encoder layer's attention (``$\mathcal{L}_{kq}^{(1)}$ only'').
In most cases, this results in considerably lower robust accuracy than using $\mathcal{L}_{ce}$ directly, indicating that for DETR ``fooling attention is all you need'' for misclassification. 




\section{Conclusion}
We revisited the robustness of transformers for image recognition against patch attacks. We identified properties of the dot-product attention's gradient that bias vanilla patch attacks to mostly ignore a core vulnerability of these attention weights. This presumably caused prior works to overestimate robustness. We propose Attention-Fool, which directly targets the dot-product attention weights and allows much tighter robustness estimates. In summary, 
Attention-Fool improves vanilla robustness evaluation across all considered vision transformers, and is able to fool DETR's global object detection with a tiny remote patch.

\noindent\textbf{Limitations.}
We focused on dot-product attention. Other attention mechanisms \cite{lee2021fnet, zhu2020deformable, wang2020linformer} 
are likely not affected to the same degree by the identified weakness. However, since dot-product attention is the predominant attention mechanism in transformers, our approach is broadly applicable.

\noindent\textbf{Potential negative societal impact.}
Adversarial attacks such as ours can be used for benign purposes like reliably evaluating robustness of ML systems as well as malign ones like exploiting weaknesses of these systems. Our research provides the basis for future work on more robust attention mechanisms that can mitigate the identified vulnerability and the resulting potential for negative societal impact.

 {\small
  \bibliographystyle{ieee_fullname}
  \bibliography{egbib}
  
 }
 
%
\newpage
\clearpage

\renewcommand{\thetable}{A\arabic{table}}
\renewcommand{\thefigure}{A\arabic{figure}}
\setcounter{table}{0}
\setcounter{figure}{0}
\setcounter{section}{0}
\setcounter{page}{1}

\appendix

\begin{center}
      \Large\textbf{Supplementary Material}
\end{center}


        \begin{table*}[t]
        \centering
        \caption{Robust accuracies (\%) under adversarial patch attacks with Attention-Fool losses when choosing different ways of aggregating $\mathcal{L}_{kq}^{hl}$ across encoder layers and attention heads. All rows are computed using \pgd$^{250}$ with momentum and step size $\alpha$=8/255. Numbers in parenthesis report the improvement or degradation in robust accuracy w.r.t. the $\smax$,$\smax$ default choice outlined in Section~\ref{sec:attention-fool}.}
        \small
        \label{tab:vit-ablation-on-reduction}
    
    \begin{tabular}{cccccccc}\toprule
            $L$ reduction & $H$ reduction & ViT-T & ViT-B & ViT-B-384 & DeiT-T & DeiT-B & DeiT-B-384
            \\\midrule
            $\smax$ & $\smax$ & 0.00 & 0.10 & 2.50 & 0.00 & 19.30 & 39.80\\
        \rowcolor{Gray} $\smax$ & mean & 0.00 \smlt{green}{(-0.00)} & 0.20 \smlt{red}{(+0.10)} & 2.70 \smlt{red}{(+0.20)} & 0.10 \smlt{red}{(+0.10)} & 17.80 \smlt{green}{(--1.50)} & 40.60 \smlt{red}{(+0.80)}\\
         $\smax$ & $\max$ & 0.00 \smlt{green}{(-0.00)} & 3.30 \smlt{red}{(+3.20)} & 5.80 \smlt{red}{(+3.30)} & 0.00 \smlt{green}{(-0.00)} & 36.60 \smlt{red}{(+17.30)} & 45.90 \smlt{red}{(+6.10)}\\
        \rowcolor{Gray}mean & $\smax$ & 0.00 \smlt{green}{(-0.00)} & 0.00 \smlt{green}{(--0.10)} & 2.90 \smlt{red}{(+0.40)} & 0.00 \smlt{green}{(-0.00)} & 18.30 \smlt{green}{(--1.00)} & 40.40 \smlt{red}{(+0.60)}\\
        mean & mean & 0.00 \smlt{green}{(-0.00)} & 0.10 \smlt{green}{(-0.00)} & 3.50 \smlt{red}{(+1.00)} & 0.00 \smlt{green}{(-0.00)} & 17.50 \smlt{green}{(--1.80)} & 39.60 \smlt{green}{(--0.20)}\\
        \rowcolor{Gray}mean & $\max$ & 0.00 \smlt{green}{(-0.00)} & 2.40 \smlt{red}{(+2.30)} & 9.20 \smlt{red}{(+6.70)} & 0.00 \smlt{green}{(-0.00)} & 23.80 \smlt{red}{(+4.50)} & 43.60 \smlt{red}{(+3.80)}\\
        $\max$ & $\smax$ & 0.10 \smlt{red}{(+0.10)} & 15.60 \smlt{red}{(+15.50)} & 26.70 \smlt{red}{(+24.20)} & 1.20 \smlt{red}{(+1.20)} & 27.10 \smlt{red}{(+7.80)} & 45.70 \smlt{red}{(+5.90)}\\
        \rowcolor{Gray}$\max$ & mean & 0.20 \smlt{red}{(+0.20)} & 13.80 \smlt{red}{(+13.70)} & 24.60 \smlt{red}{(+22.10)} & 1.50 \smlt{red}{(+1.50)} & 26.10 \smlt{red}{(+6.80)} & 46.10 \smlt{red}{(+6.30)}\\
        $\max$ & $\max$ & 3.70 \smlt{red}{(+3.70)} & 45.50 \smlt{red}{(+45.40)} & 59.70 \smlt{red}{(+57.20)} & 1.50 \smlt{red}{(+1.50)} & 58.50 \smlt{red}{(+39.20)} & 58.80 \smlt{red}{(+19.00)}\\
        \bottomrule
        \end{tabular}

    \end{table*}

\section{Gradient of Dot-Product Attention - Results}\label{sec:gradient-ratios}

        \begin{table*}[t]
        \centering
             \caption{Median of $| (\nabla_X A_h(X))  X) / (A_h(X)  1_X) |$ over tokens and heads, for models on 12 encoder layers. We report the mean of the ratio medians and its standard error over 100 randomly selected images from the MS COCO 2017 validation set~\cite{lin2014microsoft} for DETR (\textsc{dc5-r50}), and over 100 randomly selected images from the ImageNet 2012~\cite{ILSVRC15} dataset for all remaining models.}
        \footnotesize
        \label{tab:grad_ratio_full}

\begin{tabular}{lccccccc}\toprule
 & DETR & ViT-T & ViT-S & ViT-B & DeIT-T & DeIT-S & DeIT-B
\\\midrule
Layer 1 & $0.208 \pm 0.008$ & $0.079 \pm 0.001$ & $0.042 \pm 0.001$ & $0.072 \pm 0.001$ & $0.035 \pm 0.000$ & $0.045 \pm 0.001$ & $0.053 \pm 0.001$\\
\rowcolor{Gray} Layer 2 & $0.035 \pm 0.001$ & $0.040 \pm 0.001$ & $0.051 \pm 0.000$ & $0.080 \pm 0.001$ & $0.045 \pm 0.001$ & $0.045 \pm 0.000$ & $0.044 \pm 0.001$\\
Layer 3 & $0.039 \pm 0.001$ & $0.031 \pm 0.000$ & $0.027 \pm 0.000$ & $0.047 \pm 0.000$ & $0.044 \pm 0.001$ & $0.033 \pm 0.000$ & $0.042 \pm 0.000$\\
\rowcolor{Gray} Layer 4 & $0.066 \pm 0.001$ & $0.043 \pm 0.001$ & $0.028 \pm 0.000$ & $0.037 \pm 0.000$ & $0.035 \pm 0.000$ & $0.032 \pm 0.000$ & $0.057 \pm 0.000$\\
Layer 5 & $0.098 \pm 0.001$ & $0.037 \pm 0.001$ & $0.029 \pm 0.000$ & $0.033 \pm 0.000$ & $0.035 \pm 0.000$ & $0.033 \pm 0.000$ & $0.044 \pm 0.000$\\
\rowcolor{Gray} Layer 6 & $0.147 \pm 0.002$ & $0.044 \pm 0.001$ & $0.027 \pm 0.000$ & $0.044 \pm 0.001$ & $0.070 \pm 0.001$ & $0.033 \pm 0.000$ & $0.040 \pm 0.000$\\
Layer 7 & - & $0.036 \pm 0.000$ & $0.031 \pm 0.000$ & $0.040 \pm 0.000$ & $0.045 \pm 0.001$ & $0.033 \pm 0.001$ & $0.040 \pm 0.000$\\
\rowcolor{Gray} Layer 8 & - & $0.045 \pm 0.001$ & $0.046 \pm 0.001$ & $0.046 \pm 0.001$ & $0.050 \pm 0.001$ & $0.037 \pm 0.001$ & $0.039 \pm 0.000$\\
Layer 9 & - & $0.171 \pm 0.005$ & $0.087 \pm 0.002$ & $0.064 \pm 0.001$ & $0.076 \pm 0.002$ & $0.052 \pm 0.001$ & $0.049 \pm 0.001$\\
\rowcolor{Gray} Layer 10 & - & $0.308 \pm 0.006$ & $0.126 \pm 0.003$ & $0.061 \pm 0.001$ & $0.082 \pm 0.002$ & $0.077 \pm 0.002$ & $0.083 \pm 0.002$\\
Layer 11 & - & $0.428 \pm 0.009$ & $0.280 \pm 0.006$ & $0.099 \pm 0.002$ & $0.080 \pm 0.002$ & $0.100 \pm 0.003$ & $0.192 \pm 0.005$\\
\rowcolor{Gray} Layer 12 & - & $0.442 \pm 0.013$ & $0.469 \pm 0.014$ & $0.249 \pm 0.006$ & $0.212 \pm 0.008$ & $0.198 \pm 0.009$ & $0.104 \pm 0.004$\\
\bottomrule
\end{tabular}

    \end{table*}
We report an extended version of Table~\ref{tab:gradient_ratio} in
Table~\ref{tab:grad_ratio_full}, which includes up to 12 encoder layers (when available) and report the median of the gradient ratio  $| (\nabla_X A_h(X))  X) / (A_h(X)  1_X) |$, more specifically mean and its standard error of this median across 100 images.
Table~\ref{tab:grad_ratio_full} shows that gradients tend to flow along the $A_h(X)  1_X$ gradient term, in particular when back-propagating all the way up to the input (as in adversarial patch optimization).
We also find that smaller ratios are connected to less effective patches, in fact ViT-* models tend to have large ratios (in particular in later layers) compared to DeiT-*, which matches lower robust accuracies in general (see Table~\ref{tab: untargeted-main}).

\section{Effect of Input Mean on Attention Weight Robustness}~\label{app:silhouette_score}
As discussed in in Section \ref{sec:robustness_of_dot_product}, less centered inputs $X$ (larger absolute value of input mean $|\mu|$) make dot-product self-attention less robust to patch attacks on the controlled setting (when input variance is constant). We recap that the input mean is $\mathbf{\mu} = \mu \cdot \mathbf{1}$, input standard deviation $\mathbf{\sigma} = \mathbf{1}$, and $W_Q = -w \cdot \mathbb{I}_{d_k}$ and $W_K = w \cdot \mathbb{I}_{d_k}$. 

We now denote the query mean by $\mathbf{\mu}_Q = \text{Mean}(W_Q\cdot X) = -w \mathbf{\mu}$ and key mean by $\mathbf{\mu}_K = \text{Mean}(W_K\cdot X) = w \mathbf{\mu}$. We note that the element-wise distance between query and key mean is  $\vert \mathbf{\mu}^{(i)}_K - \mathbf{\mu}^{(i)}_Q  \vert = \vert w \mathbf{\mu}_i  + w \mathbf{\mu}_i \vert = 2 \vert w \vert \vert \mu \vert$.
On the other hand, for query standard deviation, we have $\mathbf{\sigma}_Q = \text{StdDev}(W_Q\cdot X) = \text{StdDev}(-w X) = w \mathbf{1}$ and for key standard deviation $\mathbf{\sigma}^2_K = \text{StdDev}(W_K\cdot X) = \text{StdDev}(w X) = w \mathbf{1}$.

As can be seen, increasing $\vert \mu \vert$ increases the distance between query and key cluster mean, while leaving the key's and query's standard deviation unchanged. As a result, it increases the separation of key and query clusters (see also Figure \ref{fig:teaser}).
On the other hand, increasing $\vert w \vert$ has no systematic effect on the separation of query and key cluster, as $w$'s effect on mean and standard deviation cancels out.

\begin{figure}[h]
	\centering
	\includegraphics[width=0.8\linewidth]{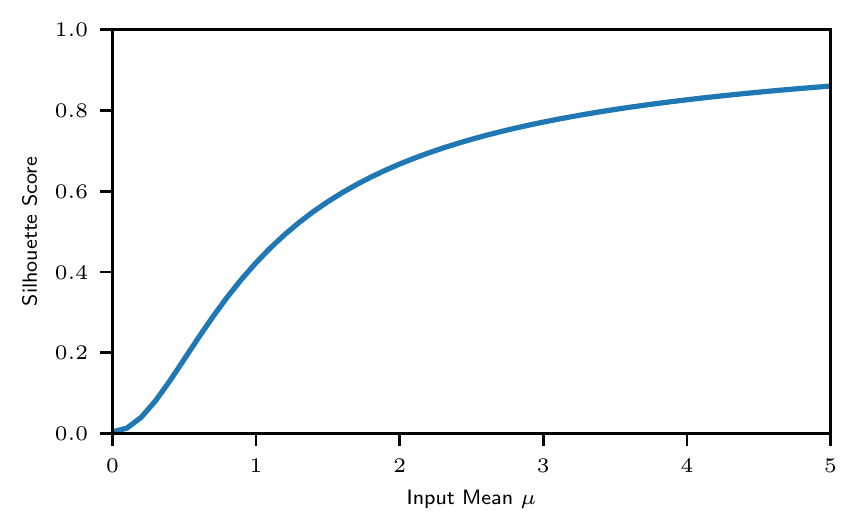}
	\caption{Silhouette score between clusters of projected keys and queries for different choices of $\mu$ on synthetic data. Larger scores corresponds to keys and queries forming more distinct clusters.}
	\label{fig:silhouette_score}
\end{figure}

We empirically quantify the query and key cluster separation using the Silhouette score \cite{Rousseeuw87silhouetteCluster}. Figure \ref{fig:silhouette_score} shows this score as a function of the input mean scale $\vert \mu \vert$ in the controlled setting. The plot confirms above's theoretical argument: increased input mean results in more separated keys and queries.

The result of this separation of keys and queries is that attention drawn by one key can increase for all queries when moving the key in the direction of the query mean (because in a sense, all queries lie in the same direction from the key as long as they are distant and have small variance). On the other hand, if keys and queries lie intermingled, than for any direction the key moves, it will get closer to some queries at the expense of increasing distance to other queries. Because of this, for an adversarial patch attack on the attention weights, it is beneficial if keys and queries are well separated (as in Figure \ref{fig:teaser}).

\section{Ablation on Attention-Fool on ViT}\label{app:lkq-vit-normalization}
In this section, we perform a series on ablations on Attention-Fool losses on ViT.
Sections~\ref{sec:vit_ablation},~\ref{app:sec:normalization-vit} and~\ref{app:lkq-vit-layer-and-head} ablate on properties of the loss, while Section~\ref{app:lkq-vit-center} and~\ref{app:sec:lkq-patch-sizes} on properties of the adversarial patch.

\subsection{Reduction over Heads and Layers}\label{sec:vit_ablation}
As discussed in Section~\ref{sec:attention-fool}, when multiple attention layers and attention heads are present, one can aggregate per-layer and per-head $\mathcal{L}^{hl}_{kq}$ in different ways.
To study the influence of these aggregation choices, in this section we empirically test three of them: (i) $\smax$ (as in Section~\ref{sec:attention-fool}), (ii) hard maximum $\max$ and (iii) mean.
Since we are aggregating along two dimensions, layers and heads, this sparks nine combinations of the three aggregations, which we empirically test on ViT models.
As shown in Table~\ref{tab:vit-ablation-on-reduction}, across  different aggregation methods, choosing $\smax$ over both layers and heads generally performs well, resulting in low robust accuracies.
Intuitively, using $\smax$ gives more flexibility to the optimization to choose the most vulnerable layers and heads.
We also find that mean aggregation across both layers and heads is an effective choice, presumably in part due to the fact that in ViT inputs to attention layers are pre-emptively normalized, leading to smaller difference in per-head per-layer behaviour.
In comparison the hard maximum $\max$ seems to be the weakest choice, leading to consistently worse results.
We note from Table~\ref{tab:vit-ablation-on-reduction} that the best combination is rather model-specific, which suggests that a similar ablation study could be run on a model-basis.
In fact, we will show in the next section how on DETR~\cite{carion2020end}, Attention-Fool with $\max$ aggregation results in better performance.

\subsection{Choosing Specific Layer}\label{app:lkq-vit-layer-and-head}

\begin{table}[tb]
\centering
     \caption{Robust accuracy (\%) of different vision transformers when choosing $l$ in $\mathcal{L}_{kq}^{l}$, which targets a specific encoder layer (see Section~\ref{sec:attention-fool}). Selecting $l$ leads to worse results compared to the baseline loss $\mathcal{L}_{kq}$.}
\footnotesize
\label{app:tab:per-layer-analysis}
\setlength{\tabcolsep}{4pt}
\begin{tabular}{r|cccc}
    \toprule 
    & ViT-T & ViT-B & DeiT-T & DeiT-B \\ \midrule
  $\mathcal{L}_{kq}$ &  0.0 & 0.1 & 0.0 & 19.3 \\ \midrule
 \rowcolor{Gray} $\mathcal{L}_{kq}^{(l)}, l=1$ &  7.5 \smlt{red}{(+7.5)}  & 0.0 \smlt{green}{(-0.1)}  & 14.0 \smlt{red}{(+14.0)}  & 36.3 \smlt{red}{(+17.0)}  \\ 
  $l=2$ &  4.4 \smlt{red}{(+4.4)}  & 39.1 \smlt{red}{(+39.0)}  & 2.2 \smlt{red}{(+2.2)}  & 56.1 \smlt{red}{(+36.8)}  \\ 
 \rowcolor{Gray} $l=3$ &  0.2 \smlt{red}{(+0.2)}  & 21.4 \smlt{red}{(+21.3)}  & 1.0 \smlt{red}{(+1.0)}  & 43.5 \smlt{red}{(+24.2)}  \\ 
  $l=4$ &  0.0 \smlt{green}{(-0.0)}  & 17.2 \smlt{red}{(+17.1)}  & 0.8 \smlt{red}{(+0.8)}  & 22.1 \smlt{red}{(+2.8)}  \\ 
 \rowcolor{Gray} $l=5$ &  0.0 \smlt{green}{(-0.0)}  & 18.8 \smlt{red}{(+18.7)}  & 1.3 \smlt{red}{(+1.3)}  & 23.9 \smlt{red}{(+4.6)}  \\ 
  $l=6$ &  0.0 \smlt{green}{(-0.0)}  & 12.4 \smlt{red}{(+12.3)}  & 0.0 \smlt{green}{(-0.0)}  & 24.6 \smlt{red}{(+5.3)}  \\ 
 \rowcolor{Gray} $l=7$ &  0.0 \smlt{green}{(-0.0)}  & 9.4 \smlt{red}{(+9.3)}  & 0.0 \smlt{green}{(-0.0)}  & 23.9 \smlt{red}{(+4.6)}  \\ 
  $l=8$ &  0.1 \smlt{red}{(+0.1)}  & 3.2 \smlt{red}{(+3.1)}  & 0.6 \smlt{red}{(+0.6)}  & 18.4 \smlt{green}{(-0.9)}  \\ 
 \rowcolor{Gray} $l=9$ &  0.0 \smlt{green}{(-0.0)}  & 0.8 \smlt{red}{(+0.7)}  & 0.1 \smlt{red}{(+0.1)}  & 21.1 \smlt{red}{(+1.8)}  \\ 
  $l=10$ &  0.1 \smlt{red}{(+0.1)}  & 1.3 \smlt{red}{(+1.2)}  & 0.1 \smlt{red}{(+0.1)}  & 23.5 \smlt{red}{(+4.2)}  \\ 
 \rowcolor{Gray} $l=11$ &  0.0 \smlt{green}{(-0.0)}  & 1.1 \smlt{red}{(+1.0)}  & 0.2 \smlt{red}{(+0.2)}  & 23.1 \smlt{red}{(+3.8)}  \\ 
  $l=12$ &  0.0 \smlt{green}{(-0.0)}  & 1.0 \smlt{red}{(+0.9)}  & 1.2 \smlt{red}{(+1.2)}  & 25.4 \smlt{red}{(+6.1)}  \\ 
    \bottomrule
\end{tabular}
\end{table}

Section~\ref{sec:attention-fool} introduces per-layer, $l$, and per-head, $h$,  losses identified by $\mathcal{L}_{kq}^{hl}$, but the evaluation of Section~\ref{sec:exp-vit} focuses on the aggregated loss $\mathcal{L}_{kq}$ and its version targeting the special token $\mathcal{L}_{kq^\star}$
To study the effectiveness of targeting specific encoder layers in the loss, here we compare using single-layer $\mathcal{L}_{kq}^{l}$ with the aggregated $\mathcal{L}_{kq}$; we report the results in Table~\ref{app:tab:per-layer-analysis}.
Targeting a single $l$ via $\mathcal{L}_{kq}^{l}$ is typically weaker than targeting all jointly via $\mathcal{L}_{kq}$.
In addition, it is unclear how to choose $l$ a priori without trying all options as there is no common pattern across models.

\begin{table}[tb]
\centering

\caption{Robust accuracy (\%) of different vision transformers for adversarial patches of different sizes  for $\mathcal{L}_{kq^\star}$. As the patch dimension shrinks to 8$\times$8 (corresponding to 0.13\% of total image pixels), the robust accuracies increase as expected.}

\small
\label{app:tab:patch-size-analysis}
\begin{tabular}{r|cccc}
    \toprule 
    Patch Size &  ViT-T & ViT-B & DeiT-T & DeiT-B \\
    \midrule
    16$\times$16 & 0.0 & 0.1 & 0.0 & 13.1\\
    \rowcolor{Gray} 14$\times$14 & 0.0 & 1.9 & 2.3 & 24.2\\
    12$\times$12 & 0.0 & 6.3 & 17.6 & 39.2\\
    \rowcolor{Gray} 10$\times$10 & 5.9 & 22.3 & 38.0 & 52.8\\
    8$\times$8 & 34.7 & 50.1 & 51.2 & 65.3\\ \midrule
    Clean Accuracy & 73.6 & 85.0 & 69.5 & 82.0 \\
    \bottomrule
\end{tabular}
\end{table}

        \begin{table*}[t]
        \centering
             \caption{Robust accuracies (\%) under Attention Fool adversarial patch attach using the un-normalized $P_Q^{hl}$ and $P_K^{hl}$ introduced in Section~\ref{sec:attention-fool}. All rows are computed using \pgd$^{250}$ with momentum and step size $\alpha$=8/255. Without normalization, $\mathcal{L}_{kq}$ and $\mathcal{L}_{kq\star}$ do not improve on $\mathcal{L}_{ce}$ baselines and in fact perform much worse, likely because individual $\mathcal{L}_{kq}^{hl}$ losses are not commensurable with each other.}
        \small
        \label{tab: no-normalization}

    \begin{tabular}{lcccccc}\toprule
            & ViT-T & ViT-B & ViT-B-384 & DeiT-T & DeiT-B & DeiT-B-384
            \\\midrule
            $\mathcal{L}_{ce}$ & 0.10 & 13.50 & 31.20 & 19.80 & 36.00 & 58.80\\
        \rowcolor{Gray}$\quad + \mathcal{L}_{kq}$ & 53.10 \smlt{red}{(+53.00)} & 81.70 \smlt{red}{(+68.20)} & 84.50 \smlt{red}{(+53.30)} & 67.40 \smlt{red}{(+47.60)} & 79.80 \smlt{red}{(+43.80)} & 80.60 \smlt{red}{(+21.80)}\\
        $\quad + \mathcal{L}_{kq*}$ & 24.40 \smlt{red}{(+24.30)} & 79.80 \smlt{red}{(+66.30)} & 82.00 \smlt{red}{(+50.80)} & 49.00 \smlt{red}{(+29.20)} & 79.60 \smlt{red}{(+43.60)} & 80.80 \smlt{red}{(+22.00)}\\
        \rowcolor{Gray}$+$ Momentum & 0.00 & 3.10 & 13.20 & 1.50 & 16.80 & 41.70\\
        $\quad + \mathcal{L}_{kq}$ & 50.00 \smlt{red}{(+50.00)} & 81.60 \smlt{red}{(+78.50)} & 84.30 \smlt{red}{(+71.10)} & 66.80 \smlt{red}{(+65.30)} & 78.30 \smlt{red}{(+61.50)} & 80.30 \smlt{red}{(+38.60)}\\
        \rowcolor{Gray}$\quad + \mathcal{L}_{kq*}$ & 21.10 \smlt{red}{(+21.10)} & 80.30 \smlt{red}{(+77.20)} & 82.10 \smlt{red}{(+68.90)} & 24.10 \smlt{red}{(+22.60)} & 78.60 \smlt{red}{(+61.80)} & 80.10 \smlt{red}{(+38.40)}\\
        \bottomrule
        \end{tabular}

    \end{table*}

        \begin{table*}[tb]
        \centering
       \caption{Robust accuracies (\%) of ResNet50 and different vision transformers 
       with the adversarial patch positioned at the center of the images. The evaluation setting is similar to Table \ref{tab: untargeted-main} except for the position of the patch. Here, the robustness of ResNet50 has further deteriorated compared to a patch placed at the image corners. In contrast, transformers demonstrate similar vulnerability with the patch positioned at the center of the images and corner of the images under our Attention Fool loss variant $\mathcal{L}_{kq\star}$. These results indicate that the transformers are less sensitive to location of the adversarial patch compared to CNNs.}
        \small
        \label{tab: untargeted-center}

     \begin{tabular}{lccccccc}\toprule
            & ResNet50 & ViT-T & ViT-B & ViT-B-384 & DeiT-T & DeiT-B & DeiT-B-384
            \\\midrule
            $\mathcal{L}_{ce}$ & 41.50 & 0.10 & 14.50 & 28.90 & 23.50 & 44.20 & 66.50\\
        \rowcolor{Gray}$\quad + \mathcal{L}_{kq} $ & -  & 0.00 \smlt{green}{(--0.10)} & 5.80 \smlt{green}{(--8.70)} & 23.10 \smlt{green}{(--5.80)} & 22.20 \smlt{green}{(--1.30)} & 44.90 \smlt{red}{(+0.70)} & 69.10 \smlt{red}{(+2.60)}\\
        $\quad + \mathcal{L}_{kq*} $ & -  & 0.00 \smlt{green}{(--0.10)} & 4.10 \smlt{green}{(--10.40)} & 18.80 \smlt{green}{(--10.10)} & 12.60 \smlt{green}{(--10.90)} & 35.90 \smlt{green}{(--8.30)} & 67.20 \smlt{red}{(+0.70)}\\
        \rowcolor{Gray}$+$ Momentum & 31.10 & 0.00 & 2.40 & 11.10 & 1.60 & 20.90 & 42.30\\
        $\quad + \mathcal{L}_{kq} $ & -  & 0.00 \smlt{green}{(-0.00)} & 0.30 \smlt{green}{(--2.10)} & 3.10 \smlt{green}{(--8.00)} & 0.10 \smlt{green}{(--1.50)} & 28.50 \smlt{red}{(+7.60)} & 48.90 \smlt{red}{(+6.60)}\\
        \rowcolor{Gray}$\quad + \mathcal{L}_{kq*} $ & -  & 0.00 \smlt{green}{(-0.00)} & 0.20 \smlt{green}{(--2.20)} & 2.60 \smlt{green}{(--8.50)} & 0.10 \smlt{green}{(--1.50)} & 11.70 \smlt{green}{(--9.20)} & 45.90 \smlt{red}{(+3.60)}\\
        \bottomrule
        \end{tabular}

    \end{table*}

        \begin{table*}[tb]
        \centering
        \caption{Adversarial patch attack success rate (\%) for a targeted attack with target class ``0''. The evaluation setting is similar to Table \ref{tab: untargeted-main}. Here, the attack success rate in vision transformers is larger than that of the ResNet50 model, but the improvements obtained with Attention Fool in comparison to the  cross-entropy baseline are not as consistent as in the untargeted attack setting.}
        \small
        \label{tab: vit-targeted}

    \begin{tabular}{lccccccc}\toprule
            & ResNet50 & ViT-T & ViT-B & ViT-B-384 & DeiT-T & DeiT-B & DeiT-B-384
            \\\midrule
            $\mathcal{L}_{ce}$ & 30.70 & 93.20 & 30.00 & 11.60 & 43.00 & 14.10 & 1.90\\
        \rowcolor{Gray}$\quad + \mathcal{L}_{kq} \quad$ & - & 94.60 \smlt{green}{(+1.40)} & 20.00 \smlt{red}{(-10.00)} & 12.20 \smlt{green}{(+0.60)} & 48.40 \smlt{green}{(+5.40)} & 14.80 \smlt{green}{(+0.70)} & 1.70 \smlt{red}{(-0.20)}\\
        $\quad + \mathcal{L}_{kq*} \quad$ & - & 94.30 \smlt{green}{(+1.10)} & 19.50 \smlt{red}{(-10.50)} & 13.00 \smlt{green}{(+1.40)} & 49.10 \smlt{green}{(+6.10)} & 14.60 \smlt{green}{(+0.50)} & 2.30 \smlt{green}{(+0.40)}\\
        \rowcolor{Gray}$+$ Momentum & 38.90 & 100.00 & 42.70 & 23.80 & 100.00 & 75.20 & 4.70\\
        $\quad + \mathcal{L}_{kq} \quad$ & - & 100.00 \smlt{green}{(+0.00)} & 43.10 \smlt{green}{(+0.40)} & 21.10 \smlt{red}{(-2.70)} & 99.10 \smlt{red}{(-0.90)} & 74.60 \smlt{red}{(-0.60)} & 4.60 \smlt{red}{(-0.10)}\\
        \rowcolor{Gray}$\quad + \mathcal{L}_{kq*} \quad$ & - & 100.00 \smlt{green}{(+0.00)} & 43.60 \smlt{green}{(+0.90)} & 22.40 \smlt{red}{(-1.40)} & 98.40 \smlt{red}{(-1.60)} & 77.10 \smlt{green}{(+1.90)} & 4.00 \smlt{red}{(-0.70)}\\
        \bottomrule
        \end{tabular}

    \end{table*}


\subsection{Normalization}\label{app:sec:normalization-vit}
In Section~\ref{sec:attention-fool} we introduced an $\ell_{1, 2}$ normalization in the computation of $\mathcal{L}_{kq}$, where projected queries and keys are normalized as $\bar{P}^{hl}_Q = \nicefrac{P^{hl}_Q}{\frac{1}{n} \vert\vert P^{hl}_Q\vert\vert_{1, 2}}$ and $\bar{P}^{hl}_K = \nicefrac{P^{hl}_K}{\frac{1}{n} \vert\vert P^{hl}_K\vert\vert_{1, 2}}$.
To evaluate the effect of this normalization we compute $\mathcal{L}_{kq}$ without it, repeating the experiment reported in Table~\ref{tab: untargeted-main}, and report the results in Table~\ref{tab: no-normalization}. We observe that this $\ell_{1, 2}$ normalization is very crucial and without normalization, performance of $\mathcal{L}_{kq}$ and $\mathcal{L}_{kq^\star}$ deteriorates considerably, to levels clearly below $\mathcal{L}_{ce}$.

\begin{table*}[tb]
	\centering
	
	\caption{Mean Average Precisions (mAP) in presence of adversarial patches computed with $\mathcal{L}_{ce}$ and Attention-Fool's $\mathcal{L}_{kq}^{(1)}$ patches on \textsc{detr} models. Differently than in Table~\ref{tab:detr-targeted4}, here we replace the adversarial target key -- one of the units in the backbone feature map -- with its clean counterpart (the same unit computed in the non-patched image).
	Replacing the adversarial key removes the entirety of the adversarial effect of Attention-Fool, showing how the mAP degradation under attack can be attributed to the target key alone.
	Three different patch sizes and four different \textsc{detr} models are considered, based either on \textsc{r50} or \textsc{r101} backbone and with and without dilation.}
	
	\footnotesize
	\label{tab:detr-replacement}
\begin{tabular}{lcccc}\toprule
        & \textsc{r50} & \textsc{dc5-r50} & \textsc{r101} & \textsc{dc5-r101}
        \\\midrule

    clean mAP & 53.00 & 54.25 & 54.41 & 56.74 \\ \midrule 
    
    \rowcolor{Gray}64$\times$64: $\mathcal{L}_{ce}$ & 38.96 & 17.51 & 33.35 & 35.67 \\ 
    $\quad + \mathcal{L}_{kq}^{(1)}$ & 51.85 \smlt{red}{(+12.90)} & 52.92 \smlt{red}{(+35.41)} & 48.85 \smlt{red}{(+15.50)} & 56.61 \smlt{red}{(+20.94)} \\ 
    \rowcolor{Gray}$\quad \mathcal{L}_{kq}^{(1)}$ only & 52.13 \smlt{red}{(+13.18)} & 53.48 \smlt{red}{(+35.97)} & 50.71 \smlt{red}{(+17.36)} & 57.11 \smlt{red}{(+21.44)} \\

    56$\times$56: $\mathcal{L}_{ce}$ & 41.18 & 25.09 & 33.64 & 38.50 \\ 
    \rowcolor{Gray}$\quad + \mathcal{L}_{kq}^{(1)}$ & 53.34 \smlt{red}{(+12.17)} & 53.66 \smlt{red}{(+28.56)} & 53.51 \smlt{red}{(+19.87)} & 56.81 \smlt{red}{(+18.31)} \\ 
    $\quad \mathcal{L}_{kq}^{(1)}$ only & 52.73 \smlt{red}{(+11.55)} & 53.28 \smlt{red}{(+28.19)} & 52.37 \smlt{red}{(+18.74)} & 57.18 \smlt{red}{(+18.68)} \\

    \rowcolor{Gray}48$\times$48: $\mathcal{L}_{ce}$ & 43.61 & 27.84 & 40.45 & 42.16 \\ 
    $\quad + \mathcal{L}_{kq}^{(1)}$ & 52.54 \smlt{red}{(+8.93)} & 52.97 \smlt{red}{(+25.13)} & 53.52 \smlt{red}{(+13.06)} & 56.90 \smlt{red}{(+14.73)} \\ 
    \rowcolor{Gray}$\quad \mathcal{L}_{kq}^{(1)}$ only & 53.27 \smlt{red}{(+9.66)} & 53.99 \smlt{red}{(+26.16)} & 53.58 \smlt{red}{(+13.13)} & 56.90 \smlt{red}{(+14.74)} \\

        \bottomrule
    \end{tabular}

\end{table*}

\subsection{Patch Location}\label{app:lkq-vit-center}
While Attention-Fool was designed to operate for arbitrary adversarial patch locations (as long as the adversarial patch aligns with ViT/DeiTs patch tiling), the location may have an effect on the resulting robust accuracies.
In Section~\ref{sec:exp-vit} we set the patch location to be the top left-most corner of the image. 
Here, we repeat the experiment but we place the patch in the image center, and we target the corresponding key. 
Specifically, in 224$\times$224-resolution models we place the patch top-left corner at 96,96, while in 384$\times$384 models we place it at 176,176.
We repeat the evaluation of Section~\ref{sec:exp-vit} in this setting, and report the results in Table~\ref{tab: untargeted-center}.
Notably, the table shows that we obtain a significant drop in ResNet50 robust accuracy because of the different location (from 49.00\% in Table~\ref{tab: untargeted-main} to 31.10\% in Table~\ref{tab: untargeted-center}), while the robust accuracies of vision transformers ViT and DeiT do not change significantly. Moreover, $\mathcal{L}_{kq^\star}$ improves upon $\mathcal{L}_{ce}$ for all but the DeiT-B-384 model.

\subsection{Patch Sizes}\label{app:sec:lkq-patch-sizes}

The evaluation of Section~\ref{sec:exp-vit} focuses on adversarial patches of size 16$\times$16, which corresponds with the token dimension in the investigated ViT models.
Here, we also investigate smaller adversarial patch sizes up to a dimension of 8$\times$8, we always place the top-left corner of the patch at the (0, 0) coordinate (top-left) of the entire image.
 We report the robust accuracies for varying sizes in Table~\ref{app:tab:patch-size-analysis}.
 As expected smaller patches lead to higher robust accuracies, but we find that even very small patches of 8$\times$8 decrease the robust accuracy significantly compared to the clean accuracy for some models.

\section{Ablation on Targeted  Attacks on ViT}\label{app:lkq-vit-targeted}
In Section~\ref{sec:exp-vit}, we reported the robust accuracies for vision transformers under an untargeted patch attacks.
Here, we also evaluate a targeted attack on the same models by selecting a target class in the optimization: rather than maximizing $\mathcal{L}_{ce}$ with the true image class $y$ as in Eq.~\ref{eq:formulation}, we replace $y$ with a target class $y^\star$ and minimize the $\mathcal{L}_{ce}$ instead (note that the $\mathcal{L}_{kq}$ loss is still maximized in a targeted setting). 
The way we combine $\mathcal{L}_{ce}$ with the Attention-Fool variants is identical as in Section~\ref{sec:exp-vit}; here we choose $y^\star = 0$.
In this case, rather than reporting robust accuracies, we report attack success rate, i.e., the percentage of times the addition of the adversarial patch changed the correct classification of an image into the target class  $y^\star $.
We report the results in Table~\ref{tab: vit-targeted}, note that the colors and the signs of improvements/degradation are inverted in comparison to the other tables.
Table~\ref{tab: vit-targeted} shows that the benefit of Attention-Fool in targeted attacks is less clear. 
We hypothesise that different weighting of $\mathcal{L}_{ce}$ and $\mathcal{L}_{kq}$ might be required in targeted settings, specifically because $\mathcal{L}_{ce}$ is bounded by zero when minimized (while it is unbounded when maximized in the untargeted setting).
Additionally, specific properties of the target class ``0'' on certain models might bias the evaluation; a larger, more in-depth evaluation of targeted attacks is left for future work.

\section{Adversarial Token Replacement}
\label{app:adv-key-replacement}

In Section~\ref{sec:exp-detr} we showed how targeting a single key with Attention-Fool in \textsc{DETR} can lead to large degradation in mAP.
Differently than in ViT models, because of DETR's hybrid architecture (CNN plus Transformer) an adversarial patch placed on the DETR image input affects a number of tokens in the encoder's input: all those backbone outputs (tokens) whose receptive field via the CNN overlaps with the input image patch.
Here, to show that the attack performance can be attributed to the individual key token that is the Attention-Fool target, we replace this key with its clean counterpart. 
To do so, we compute all keys on the clean image and all keys in the patched image (by forwarding the image through the CNN backbone), and we replace the target adversarial key with its clean counterpart.
This is either the key indexed by 2,2 or by 4,4 in non-dilated and dilated models, respectively.
We report the resulting mAPs in Table~\ref{tab:detr-replacement}.
The table shows how replacing this single token's key removes a large part of the adversarial effect in all rows using $\mathcal{L}_{kq}^{(1)}$ loss.
In comparison, in rows using $\mathcal{L}_{ce}$, replacing this token's key still results in low mAPs, showing that in this case the adversarial effect is not attributable to the same adversarial token alone.

\section{Additional Visualizations}\label{sec:additional-visualizations.}
\begin{figure*}[t]
     \centering
     \begin{subfigure}[b]{0.99\textwidth}
         \centering
        \includegraphics[width=\textwidth]{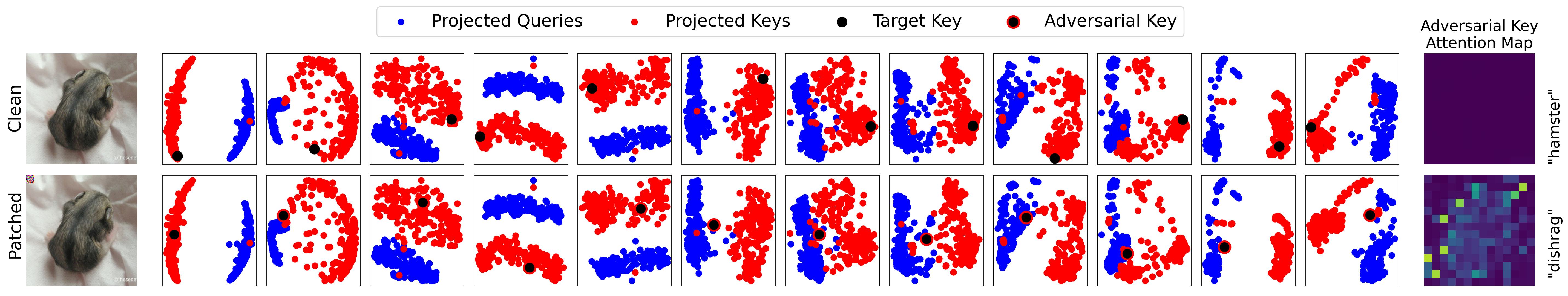}
     \end{subfigure}
     \hfill
     \begin{subfigure}[b]{0.99\textwidth}
         \centering
         \includegraphics[width=\textwidth]{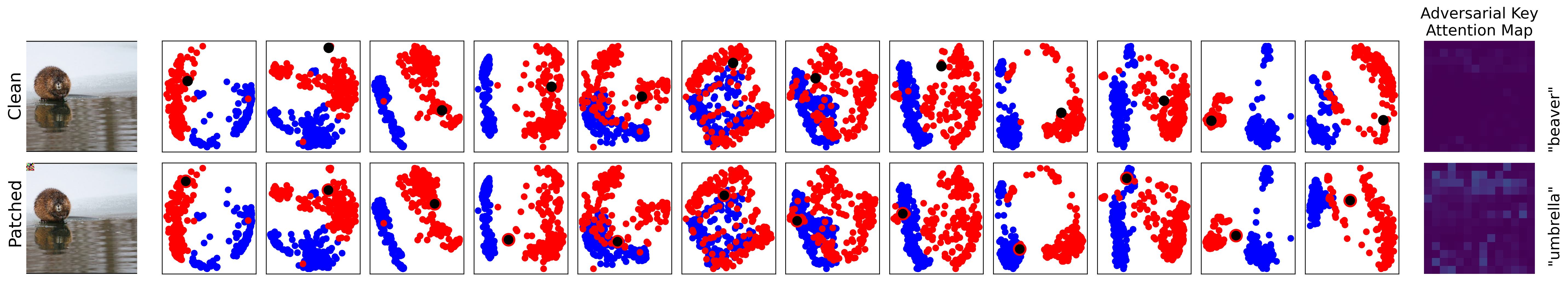}
     \end{subfigure}
     \hfill
     \begin{subfigure}[b]{0.99\textwidth}
         \centering
         \includegraphics[width=\textwidth]{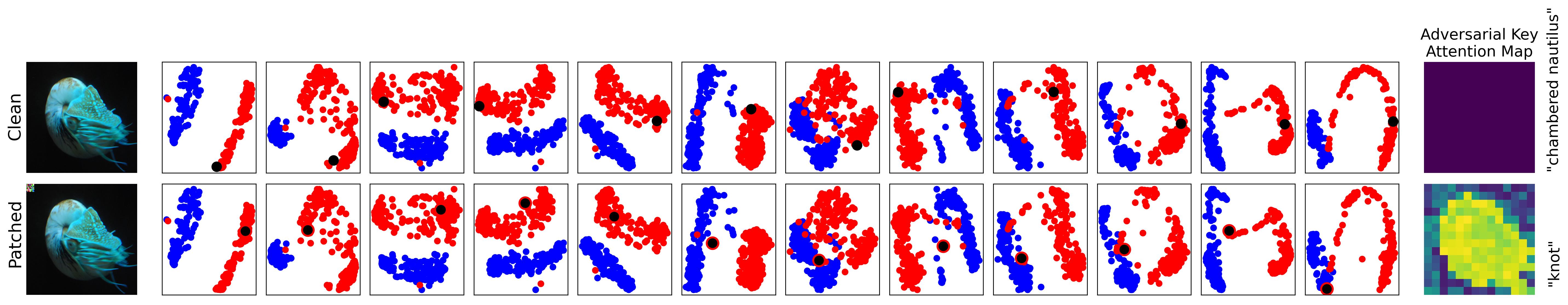}
     \end{subfigure}
     \hfill
     \begin{subfigure}[b]{0.99\textwidth}
         \centering
         \includegraphics[width=\textwidth]{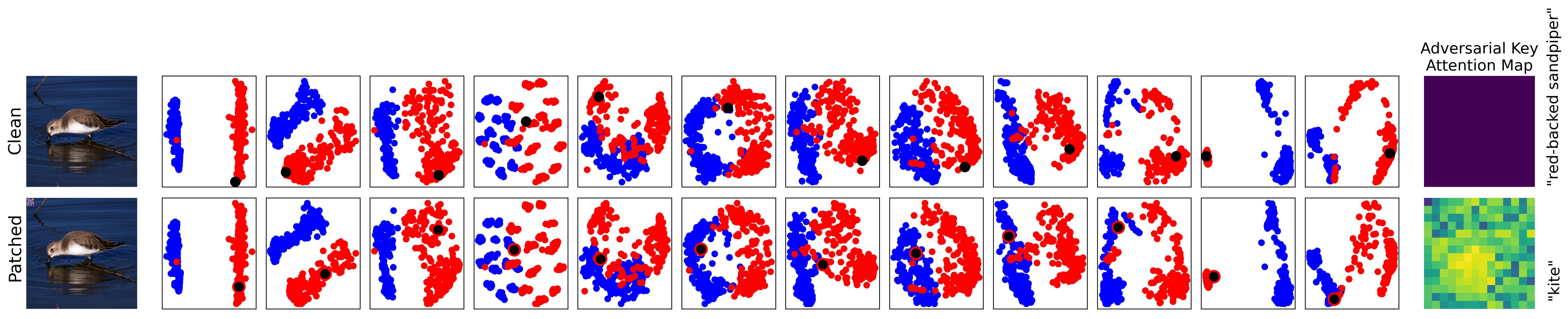}
     \end{subfigure}
     \hfill
     \begin{subfigure}[b]{0.99\textwidth}
         \centering
         \includegraphics[width=\textwidth]{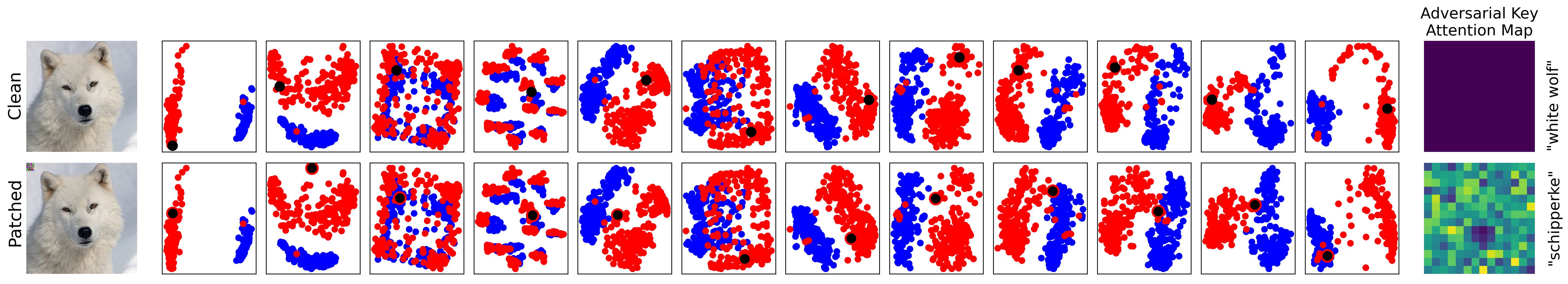}
     \end{subfigure}
     \hfill
     \begin{subfigure}[b]{0.99\textwidth}
         \centering
         \includegraphics[width=\textwidth]{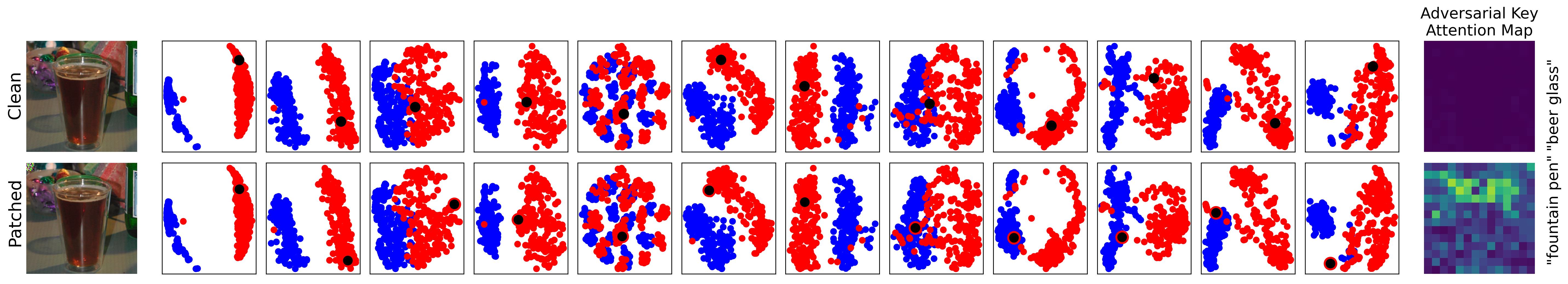}
     \end{subfigure}
    \caption{Embedded projected key and query tokens for clean and patched input images on each of the 12 DeiT-B layers, for a single attention head. For each image we chose an attention head which showed large amount of changes. The last column reports the attention map weights of the adversarial key on the last layer -- showing that generally the key draws a large amount of attention from queries. Note that while the adversarial patch tends to focus on drawing the attention on the last layer (which is visualized in the right-most column), it can target arbitrary layers -- in fact, we obtain successful mis-classifications even if the last layer attention map only has minor changes.}
    \label{fig:more-per-layer}
\end{figure*}

\begin{figure*}[t]
     \centering
     \begin{subfigure}[b]{0.63\textwidth}
         \centering
        \includegraphics[width=\textwidth]{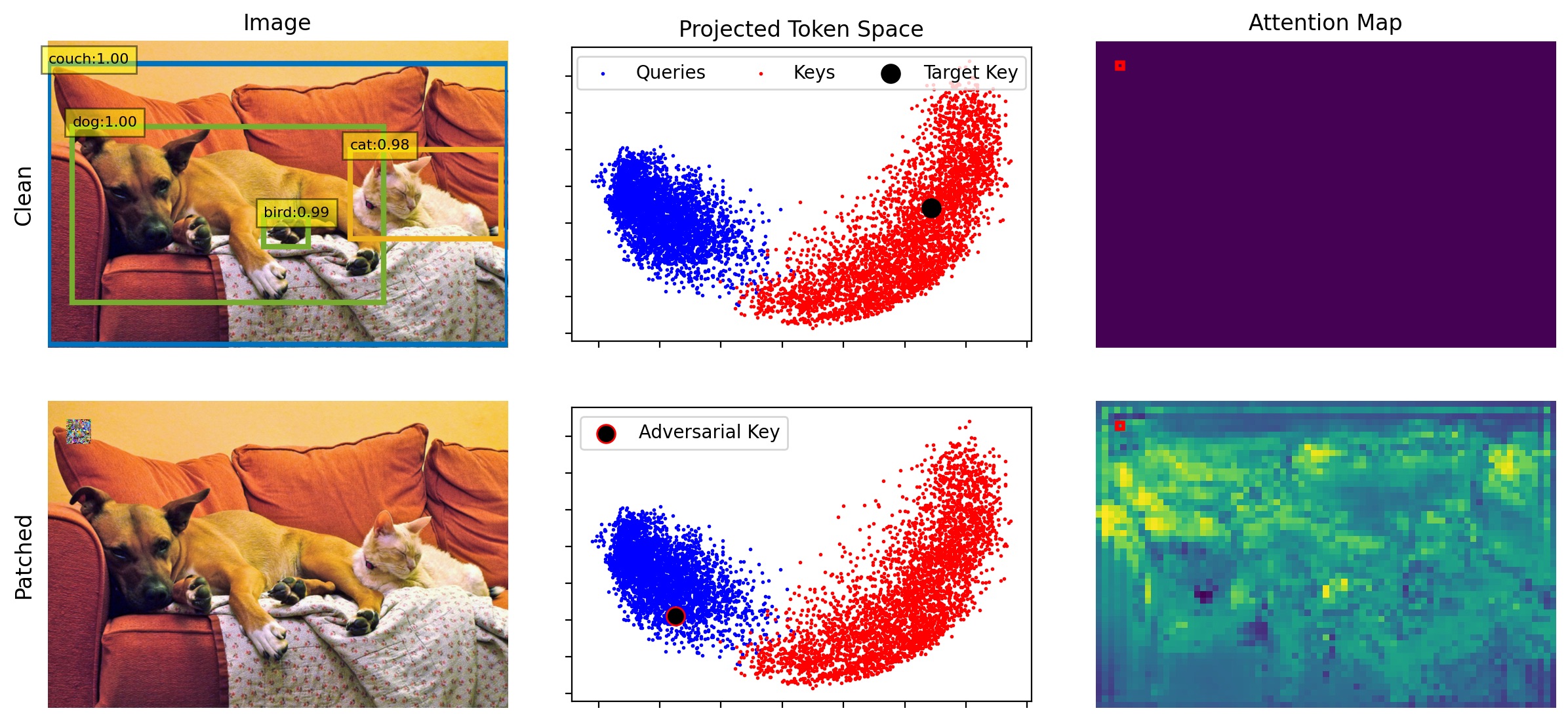}
     \end{subfigure}
     \hfill
     \begin{subfigure}[b]{0.63\textwidth}
         \centering
         \includegraphics[width=\textwidth]{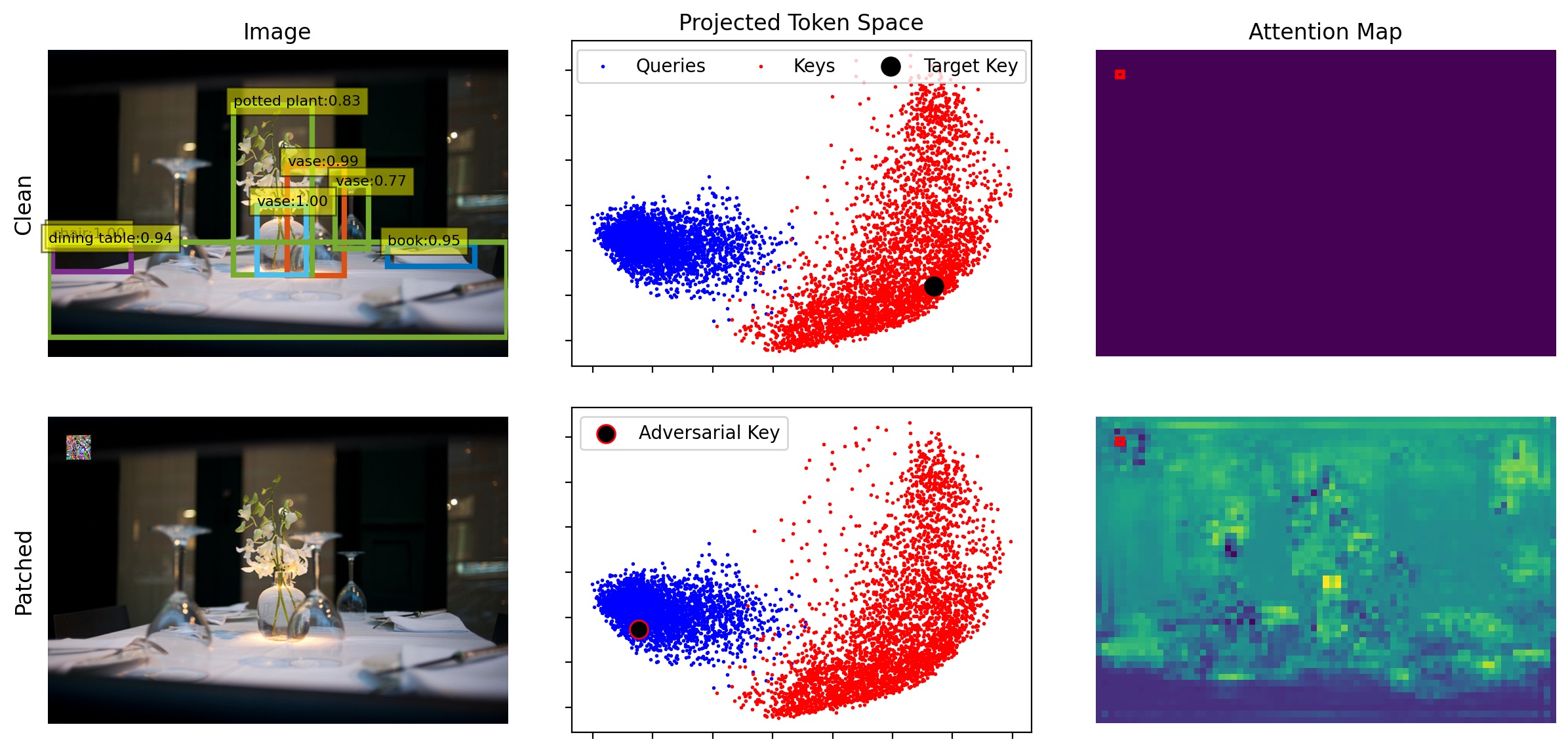}
     \end{subfigure}
     \hfill
     \begin{subfigure}[b]{0.63\textwidth}
         \centering
         \includegraphics[width=\textwidth]{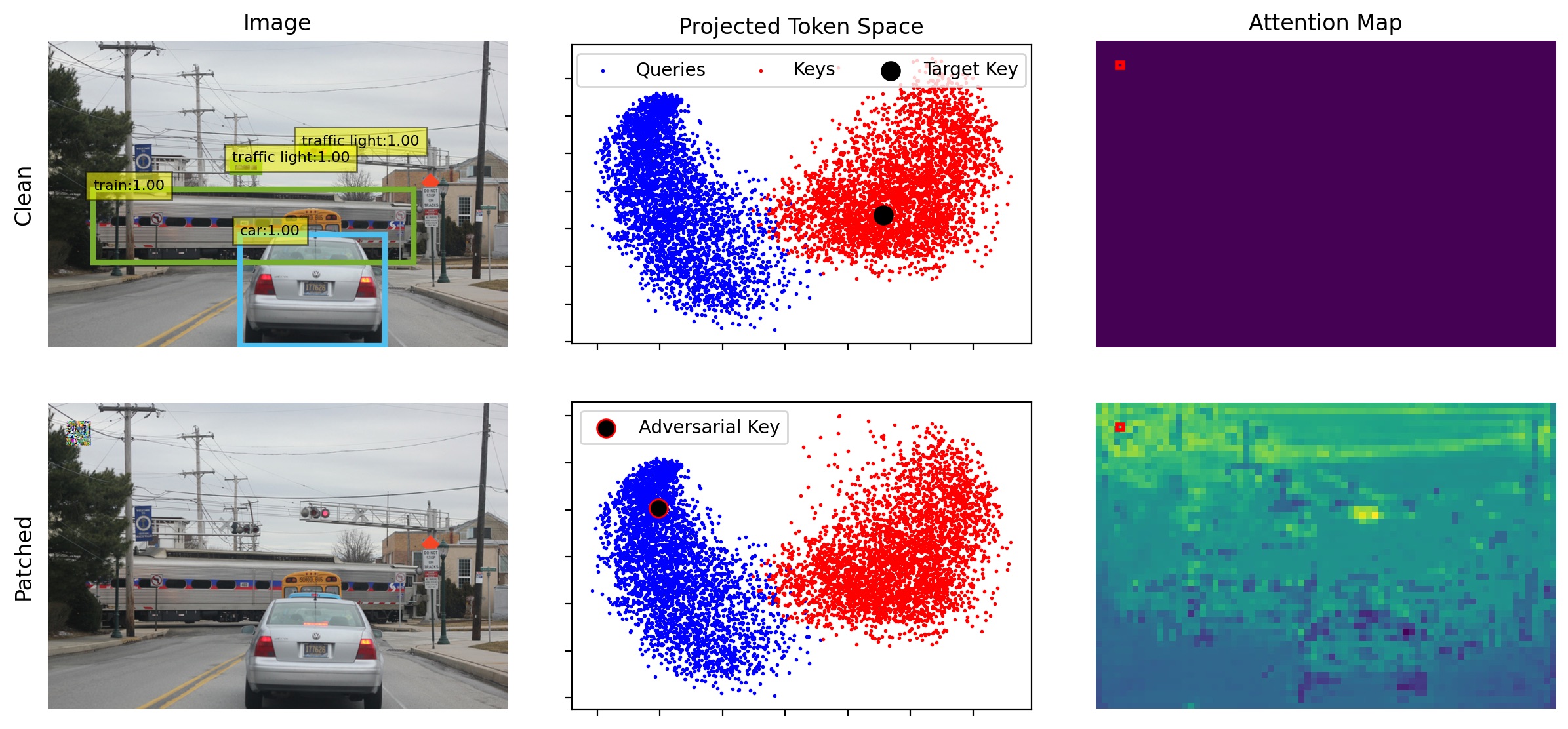}
     \end{subfigure}
     \hfill
    \begin{subfigure}[b]{0.63\textwidth}
        \centering
        \includegraphics[width=\textwidth]{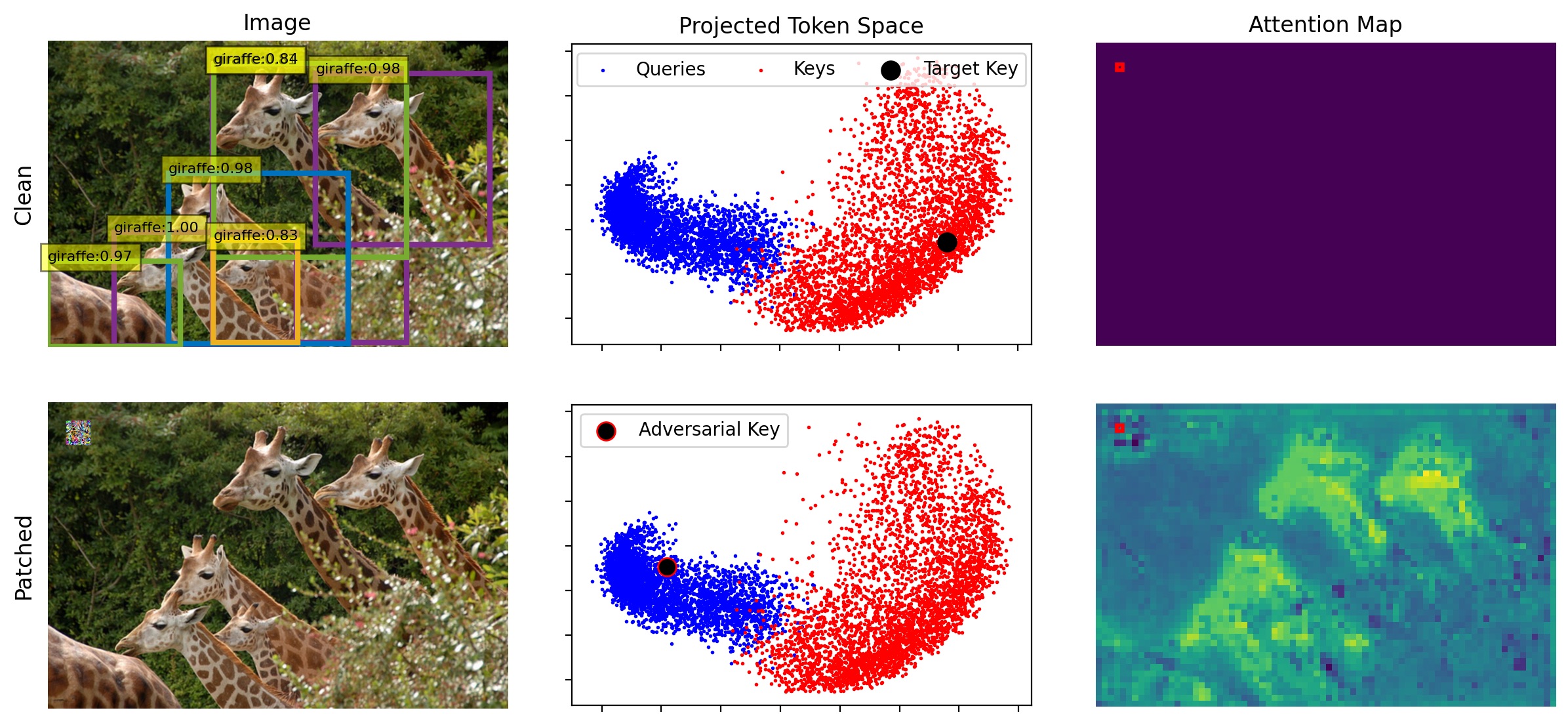}
    \end{subfigure}
    \caption{More comparisons of clean and adversarially patched input for DETR~\cite{carion2020end}. The patch shifts a targeted key token towards the cluster of query tokens. In dot-product attention, this directs queries attention to the malicious token and prevents the model from detecting the remaining objects. The right-most column compares queries' attention weights to the adversarial key, whose location is marked by a red box, between clean and patched inputs. These images use \textsc{detr dc5-r50} and patches are optimized with $\mathcal{L}_{ce} + \mathcal{L}_{kq}^{(1)}$.}
    \label{fig:more-teaser}
\end{figure*}

We report additional visualizations akin to those in Fig.~\ref{fig:teaser} and Fig.~\ref{fig:vit-per-layer-visualization} in Figure~\ref{fig:more-teaser} and Figure~\ref{fig:more-per-layer}, respectively.

\end{document}